\documentclass{article}

 \usepackage[preprint]{neurips_2026}

\usepackage[utf8]{inputenc} % allow utf-8 input
\usepackage[T1]{fontenc}    % use 8-bit T1 fonts
\usepackage{hyperref}       % hyperlinks
\usepackage{url}            % simple URL typesetting
\usepackage{booktabs}       % professional-quality tables
\usepackage{amsfonts}       % blackboard math symbols
\usepackage{nicefrac}       % compact symbols for 1/2, etc.
\usepackage{microtype}      % microtypography
\usepackage{xcolor}         % colors
\usepackage[table]{xcolor}
\usepackage{svg}
\usepackage{ulem}
\usepackage{graphicx}
\usepackage{subcaption}
\usepackage[most]{tcolorbox}
\usepackage{listings}
\usepackage{arydshln}
\definecolor{codegray}{rgb}{0.5,0.5,0.5}
\definecolor{backcolour}{rgb}{0.95,0.95,0.92}

\usepackage{enumitem}
\usepackage{fontawesome5}
\usepackage{caption}

\definecolor{originalcolor}{RGB}{52, 152, 219}   % Blue
\definecolor{geminicolor}{RGB}{155, 89, 182}     % Purple
\definecolor{qwencolor}{RGB}{230, 126, 34}       % Orange
\definecolor{revisedcolor}{RGB}{39, 174, 96}     % Green
\definecolor{finalcolor}{RGB}{231, 76, 60}       % Red
\definecolor{lightgray}{RGB}{245, 245, 245}

\hypersetup{
    colorlinks=true,
    linkcolor=blue,
    urlcolor=blue
}

\newtcolorbox{originalbox}[1][]{
    enhanced,
    colback=originalcolor!5,
    colframe=originalcolor,
    boxrule=1pt,
    arc=3mm,
    title=#1,
    fonttitle=\bfseries\color{white},
    coltitle=white,
    attach boxed title to top left={xshift=5mm, yshift=-3mm},
    boxed title style={colback=originalcolor, sharp corners}
}

\newtcolorbox{geminibox}[1][]{
    enhanced,
    colback=geminicolor!5,
    colframe=geminicolor,
    boxrule=1pt,
    arc=3mm,
    title=#1,
    fonttitle=\bfseries\color{white},
    coltitle=white,
    attach boxed title to top left={xshift=5mm, yshift=-3mm},
    boxed title style={colback=geminicolor, sharp corners}
}

\newtcolorbox{qwenbox}[1][]{
    enhanced,
    colback=qwencolor!5,
    colframe=qwencolor,
    boxrule=1pt,
    arc=3mm,
    title=#1,
    fonttitle=\bfseries\color{white},
    coltitle=white,
    attach boxed title to top left={xshift=5mm, yshift=-3mm},
    boxed title style={colback=qwencolor, sharp corners}
}

\newtcolorbox{revisedbox}[1][]{
    enhanced,
    colback=revisedcolor!5,
    colframe=revisedcolor,
    boxrule=1pt,
    arc=3mm,
    title=#1,
    fonttitle=\bfseries\color{white},
    coltitle=white,
    attach boxed title to top left={xshift=5mm, yshift=-3mm},
    boxed title style={colback=revisedcolor, sharp corners}
}

\newtcolorbox{finalbox}[1][]{
    enhanced,
    colback=finalcolor!5,
    colframe=finalcolor,
    boxrule=1pt,
    arc=3mm,
    title=#1,
    fonttitle=\bfseries\color{white},
    coltitle=white,
    attach boxed title to top left={xshift=5mm, yshift=-3mm},
    boxed title style={colback=finalcolor, sharp corners}
}

\newcommand{\scorebadge}[2]{%
    \tcbox[on line, colback=#1, colframe=#1, coltext=white,
           boxrule=0pt, arc=2mm, left=4pt, right=4pt, top=2pt, bottom=2pt]{\bfseries Score: #2/10}%
}

\title{ERQA-Plus: A Diagnostic Benchmark for Reasoning in Embodied AI}
\author{%
  Hong~Yang\\
  Centre for Frontier AI Research\\
  Agency for Science, Technology and Research\\
  Singapore\\
  \texttt{Yang\_Hong@a-star.edu.sg} \\
  \And
  Basura Fernando \\
  Centre for Frontier AI Research,
  Agency for Science, Technology and Research,  Singapore  \\
  College of Computing and Data Science, Nanyang Technological University, Singapore
}

\begin{document}

\maketitle

\begin{abstract}
Generalist embodied agents require more than object recognition: they must reason about spatial relations, actions, procedures, human intentions, environmental constraints, and commonsense consequences from situated visual observations. Yet existing visual and embodied question answering benchmarks often provide limited control over the reasoning dependencies being tested, making it difficult to distinguish grounded embodied reasoning from shortcut-driven visual or linguistic pattern matching. We present \textbf{ERQA-Plus}, a diagnostic benchmark for reasoning in embodied AI. ERQA-Plus contains 1,766 question-answer instances grounded in 711 robot-centric images and organized according to a structured taxonomy spanning perceptual, action-centric, social-interaction, navigation-environmental, and contextual commonsense reasoning. The dataset is constructed using a multi-stage generation and validation pipeline that combines taxonomy-guided question generation, automatic quality judging, iterative revision, and human assessment to improve visual grounding, answer validity, and reasoning quality. We benchmark representative general-purpose vision-language models and embodied models, including LLaVA-NeXT-8B, Prismatic-7B, MiniCPM-V-4.5-8B, Qwen3-VL, RoboRefer-8B, and RoboBrain2.5-8B. Although the strongest model, Qwen3-VL-32B, achieves 83.4\% overall accuracy and 61.4 SBERT score, category-level results reveal persistent weaknesses in spatial reasoning, procedural reasoning, event prediction, and intention inference. ERQA-Plus therefore provides a fine-grained evaluation framework for measuring not only whether embodied agents answer correctly, but also which forms of embodied reasoning they can and cannot perform reliably.
The dataset is available \url{https://huggingface.co/datasets/huggingdas/erqa-plus} and the project page at~ \url{https://github.com/LUNAProject22/erqa-plus}.
\end{abstract}

\section{Introduction}
\begin{comment}
Motivation: Why embodied QA remains under-evaluated\\
Limitations of existing VQA and EQA benchmarks\\
Gap: Lack of controlled reasoning dependency, \sout{difficulty calibration}, and validation\\
Contributions:\\
A structured embodied reasoning benchmark (ERQA-plus)\\
Multi-stage QA generation-validation pipeline\\
\sout{Difficulty-aware evaluation protocol}\\
Human-validated quality scoring\\
Comprehensive benchmarking across general VLMs and embodied agents\\
\end{comment}

Embodied AI systems aim to integrate perception, reasoning, and action to operate effectively in dynamic, real-world environments~\cite{zhong2025survey}. Recent advances in vision--language models (VLMs) and embodied agents have demonstrated promising capabilities in visual understanding and instruction following~\cite{chi2025diffusion,team2024octo,reuss2025flower}. However, despite these advances, evaluating whether such systems truly perform grounded reasoning as opposed to correlations remains a significant challenge.

Existing benchmarks in Visual Question Answering~\cite{yue2024mmmu,chen2024we,li2024mvbench} (VQA) and Embodied Question Answering~\cite{majumdar2024openeqa} (EQA) provide useful testbeds but exhibit critical limitations. Many VQA datasets are known to suffer from shortcut learning, linguistic priors, and weak grounding requirements, allowing models to achieve high performance without deep visual reasoning. Meanwhile, EQA benchmarks often emphasize navigation or task completion but provide limited control over the underlying reasoning processes being evaluated. As a result, current benchmarks do not adequately disentangle different reasoning skills nor provide fine-grained diagnostics of model capabilities.
This gap is particularly problematic for generalist embodied agents, which must coordinate multiple forms of reasoning including perception, action understanding, spatial awareness, and commonsense inference within a unified decision-making process. Without structured and controlled evaluation, it is difficult to identify failure modes, measure progress, or guide the development of more robust and reliable embodied intelligence.

To address these challenges, we introduce \textbf{ERQA-Plus}, a diagnostic benchmark designed to systematically evaluate reasoning in embodied AI. Unlike prior datasets, ERQA-Plus emphasizes controlled reasoning dependencies, strong visual grounding, and rigorous validation. The benchmark is constructed through a multi-stage pipeline that integrates automated generation, structured validation, and iterative refinement, ensuring high-quality question-answer pairs that require genuine embodied reasoning.
Specifically, ERQA-Plus makes the following contributions:
\begin{itemize}
    \item \textbf{A structured embodied reasoning benchmark} that organizes evaluation into a taxonomy of interdependent reasoning categories, enabling fine-grained analysis of model capabilities.
    \item \textbf{A multi-stage QA generation and validation pipeline} that produces visually grounded, high-quality questions with minimal ambiguity and reduced linguistic bias.
    \item \textbf{A quality-controlled evaluation framework} incorporating automated judging and human validation to ensure reliability and benchmarking rigor.
    \item \textbf{Comprehensive benchmarking} across both general-purpose VLMs and embodied agents, providing insights into their strengths and limitations in grounded reasoning.
\end{itemize}

Through ERQA-Plus, we aim to move beyond coarse accuracy metrics and toward a more diagnostic evaluation paradigm that reveals how and why embodied AI systems succeed or fail. We believe this benchmark will facilitate the development of more robust, interpretable, and cognitively grounded embodied intelligence.

\section{Dataset creation}

\begin{figure}[t]
    \centering
    \includegraphics[width=1\textwidth]{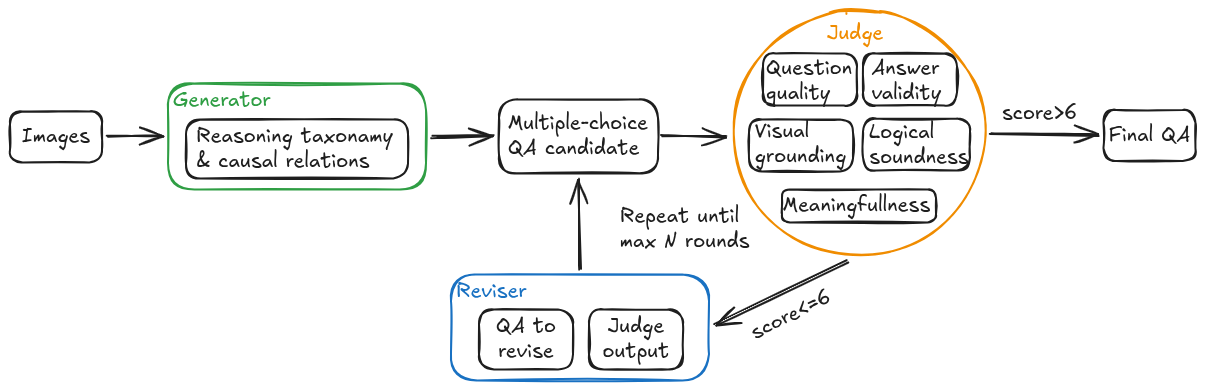}
    \caption{    
    Dataset creation pipeline. The question generation pipeline consists of three agents 1) the Generator 2) the Judge and 3) Reviser. Curation pipeline is applied iteratively over Judge and Reviser agents for maximum of N rounds to generate high quality visually grounded embodied AI questions.
    }
    \label{fig:dataset_creation}
\end{figure}

\subsection{Reasoning taxonomy for generalist embodied agents}
Generalist embodied agents operate in open-ended environments that require tight coupling between perception, action, language, and social interaction. To systematically evaluate such capabilities, we organize reasoning into five interdependent categories that reflect the cognitive architecture required for navigation, manipulation, and human collaboration. This taxonomy extends beyond static visual understanding and captures the multi-stage, goal-directed reasoning processes characteristic of embodied intelligence.

\paragraph{Perceptual reasoning} Perceptual reasoning constitutes the foundation of embodied cognition. It focuses on recognizing entities in the environment and modeling their spatial and temporal relationships. This category includes: (i) \textit{Object reasoning}, which involves identifying objects and inferring their physical or functional attributes; (ii) \textit{Spatial reasoning}, which captures geometric structure, relative positioning, containment, support, and occlusion; and (iii) \textit{Temporal reasoning}, which models event sequences, motion dynamics, and short-term causality. Robust perceptual representations are essential, as they ground all higher-level reasoning processes.

\paragraph{Action-centric reasoning} Beyond perception, embodied agents must interpret and generate behavior. Action-centric reasoning addresses the understanding of actions, goals, and task structures. It comprises: (i) \textit{Action recognition}, which identifies ongoing manipulations or movements; (ii) \textit{Intention reasoning}, which infers underlying goals and predicts future actions from partial observations; and (iii) \textit{Procedural reasoning}, which models multi-step tasks, determines valid action sequences, and assesses task completion. This category forms the bridge between perception and control, enabling the translation of scene understanding into executable plans.

\paragraph{Social and interaction reasoning} In collaborative environments, agents must interpret human mental states and communicative cues. Social and interaction reasoning supports effective and safe collaboration by enabling \textit{Attention and role understanding}, including the inference of gaze, task roles, leadership, and division of labor. These capabilities are critical for shared autonomy and human-in-the-loop systems.

\paragraph{Navigation and environmental reasoning} This category focuses on understanding scene layouts and acting within them. It includes: (i) \textit{Layout understanding}, which models scene structure and object–environment relationships; and (ii) \textit{Path planning}, which determines feasible trajectories while avoiding obstacles and respecting environmental constraints. This reasoning type tightly couples spatial understanding with action feasibility.

\paragraph{Contextual and commonsense reasoning}
Real-world environments are inherently ambiguous and incomplete. Contextual and commonsense reasoning enables agents to leverage prior knowledge and functional understanding to interpret scenes and anticipate outcomes. It includes: (i) \textit{Functional context}, which infers the intended use of objects or spaces; (ii) \textit{Plausible event prediction}, which anticipates likely future states and causal consequences; and (iii) \textit{World knowledge grounding}, which applies commonsense priors to resolve uncertainty. This category interacts bidirectionally with all others by constraining interpretations and guiding decision-making.

\paragraph{Interdependency structure}
The proposed taxonomy reflects a hierarchical and interdependent cognitive structure rather than a flat categorization. Perceptual reasoning underpins action-centric, navigation, and social reasoning. Action-centric reasoning relies on perceptual grounding and environmental constraints. Social reasoning informs intention inference and collaborative planning. Contextual and commonsense reasoning provides global priors that influence all reasoning stages.

Overall, this decomposition enables fine-grained evaluation and diagnostic analysis of embodied intelligence. It distinguishes systems that perform isolated visual recognition from those that integrate perception, action, context, and social understanding into coherent, goal-directed behavior.

\subsection{Generator}
The question generator transforms robot-centric visual observations into structured multiple-choice questions that probe grounded reasoning capabilities. Given one or more images captured from a robotic gripper’s main or side camera, the generator produces 2–5 high-quality questions aligned with a predefined embodied reasoning taxonomy. Each question targets a specific reasoning subtype and is designed to require visual grounding rather than simple pattern recognition.

\paragraph{Design principles} The generator is designed to prioritize grounded visual reasoning over linguistic shortcuts, ensuring that questions require evidence-based interpretation of the scene rather than reliance on language priors. Each question is explicitly aligned with a well-defined reasoning subtype, maintaining clarity and consistency within the taxonomy. The sequence of questions preserves causal coherence, reflecting realistic embodied cognition pipelines from perception to action and higher-level inference. All content remains relevant to embodied robotic tasks such as navigation, manipulation, and collaboration. Additionally, answer choices are carefully constructed with balanced and plausible distractors to encourage robust reasoning and minimize superficial guessing.

\paragraph{Taxonomy-guided question construction} The generator operates under a hierarchical reasoning framework consisting of five major categories: Perceptual, Action-Centric, Social and Interaction, Navigation and Environmental, and Contextual and Commonsense Reasoning. Each question is explicitly labeled with one fine-grained subtype. To ensure structural coherence, the generator respects causal dependencies among reasoning types. Questions are sequenced such that later questions may implicitly rely on earlier perceptual understanding, mimicking real embodied decision pipelines.

% \paragraph{Visual grounding constraints} To ensure the benchmark maintains high standards of scientific rigor and practical utility for embodied AI, the formulation of each question is governed by several core constraints. Specifically, inquiries are constructed to avoid revealing explicit visual cues within the text, thereby forcing the model to rely entirely on the visual input. These questions require sophisticated reasoning grounded in observable scene elements, moving beyond trivial object recognition. Instead of mere listing, the benchmark emphasizes spatial, functional, and procedural understanding—capabilities that are critical for agents interacting with physical environments.

% The design of the answer options follows a structured methodology to mitigate shortcut learning and linguistic bias. Each set of options consists of plausible, mutually exclusive candidates that are specifically engineered to be visually confusable if only shallow reasoning is applied. To prevent the model from guessing based on statistical language patterns, the distractors are balanced to minimize the influence of linguistic priors. Every question includes exactly one correct answer, which is supported by a concise visual justification to ensure interpretability and verifiable grounding.

\paragraph{Visual grounding constraints} Each question is formulated to withhold explicit visual cues from the text, compelling models to reason directly from scene observations rather than linguistic shortcuts. Questions target spatial, functional, and procedural understanding rather than trivial object recognition, reflecting the reasoning demands of embodied AI agents. Answer options are designed to be plausible, mutually exclusive, and visually confusable under shallow reasoning, with distractors balanced to suppress linguistic priors. Each question has exactly one correct answer accompanied by a concise visual justification to ensure interpretability and verifiable grounding.

\subsection{Judge agent}
To ensure the reliability and benchmarking value of ERQA-Plus, we introduce a Judge agent that automatically evaluates the quality of each embodied visual question-answer (QA) instance. The Judge agent acts as a meta-evaluator, verifying whether a QA sample is visually grounded, logically coherent, spatially correct, and suitable for training or benchmarking embodied reasoning models.

\paragraph{Role and objective} Given image(s), a multiple-choice question, answer options, and ground-truth label, it assesses whether the QA instance is: (1) meaningful and grounded in visual evidence; (2) logically and spatially sound; (3) unambiguous in its answer choices, and (4) correctly labeled. Its output is restricted to a structured JSON object containing a numerical score, a concise justification, and a list of detected issues. This design enforces deterministic, machine-readable evaluation for scalable dataset curation.

\paragraph{Evaluation dimensions} Each QA instance is assessed along five complementary dimensions:

\textbf{Question quality.} The agent evaluates whether the question is directly grounded in the provided image(s), reflects accurate visual perception, encodes valid spatial relationships (e.g., relative position, distance, and orientation), and requires non-trivial embodied reasoning rather than superficial observation.

% \textbf{Option quality.} The agent verifies that the answer choices are mutually exclusive, visually distinguishable, and collectively well-formed, with exactly one correct option and no missing, redundant, or ambiguously phrased alternatives.

% \textbf{Ground-truth validity.} The agent ensures that the annotated ground-truth answer is fully supported by the visual evidence. Any discrepancy between the image content and the labeled answer is treated as a critical error.

\textbf{Option quality and ground-truth validity.} The agent ensures that answer choices are mutually exclusive, visually distinguishable, and well-formed with exactly one correct option, while verifying that the ground-truth answer is fully supported by the visual evidence and free of discrepancies.

\textbf{Spatial reasoning correctness.} For spatially grounded questions, the agent evaluates the correctness of reasoning involving distance comparisons, occlusion handling, viewpoint-dependent interpretation, and object–object spatial relationships.

\textbf{Overall usefulness.} Finally, the agent assesses whether the QA instance provides meaningful training signal and contributes to robust benchmarking of embodied reasoning capabilities.

\paragraph{Scoring protocol} The Judge Agent assigns a quality score in the range [0, 10] according to the following rubric: scores of 9–10 indicate high-quality, well-grounded, and unambiguous instances with strong training value; 7–8 correspond to generally sound samples with minor imperfections; 5–6 reflect noticeable deficiencies, such as uncertain spatial reasoning or questionable annotations; 3–4 denote substantial issues, including ambiguity or incorrect labeling; and 0–2 represent fundamentally flawed instances that are visually inconsistent or unusable. This fine-grained scoring scheme facilitates automatic filtering, curriculum design, and the construction of quality-controlled benchmark splits.

\subsection{Reviser agent}
% To ensure high-quality, reliable, and pedagogically valuable QA pairs in ERQA-Plus, we introduce a Reviser Agent that refines candidate QAs based on structured feedback from the Judge Agent. While the Question Generator produces visually grounded multiple-choice questions and the Judge evaluates their quality, the Reviser serves as an iterative correction module that improves clarity, grounding, and reasoning rigor before inclusion in the benchmark.

% The Reviser Agent functions as a quality amplification mechanism in our iterative data curation loop. Empirically, this stage substantially reduces ambiguity, eliminates weak distractors, and increases reasoning depth—particularly in spatial reasoning, functional context understanding, and procedural inference. By incorporating structured feedback before finalization, ERQA-Plus achieves higher consistency, stronger supervision signals, and improved benchmarking reliability for generalist embodied agents.

We introduce a Reviser agent to refine candidate QA pairs using structured feedback from the Judge agent. Acting as an iterative correction module within the data curation loop, it improves clarity, grounding, and reasoning while reducing ambiguity and eliminating weak distractors. This process enhances reasoning depth—especially in spatial, functional, and procedural contexts—and yields more reliable benchmarking.

Overall our high-quality question generation pipeline consists of three components 1) the Generator 2) the Judge agent and 3) Reviser agent. The overall pipeline is applied iteratively over Judge agent and Reviser agent for a generated question as shown in Figure~\ref{fig:dataset_creation}.

\subsection{Dataset statistics and analysis}
\begin{figure}[b]
    \centering
    \includegraphics[width=0.6\linewidth]{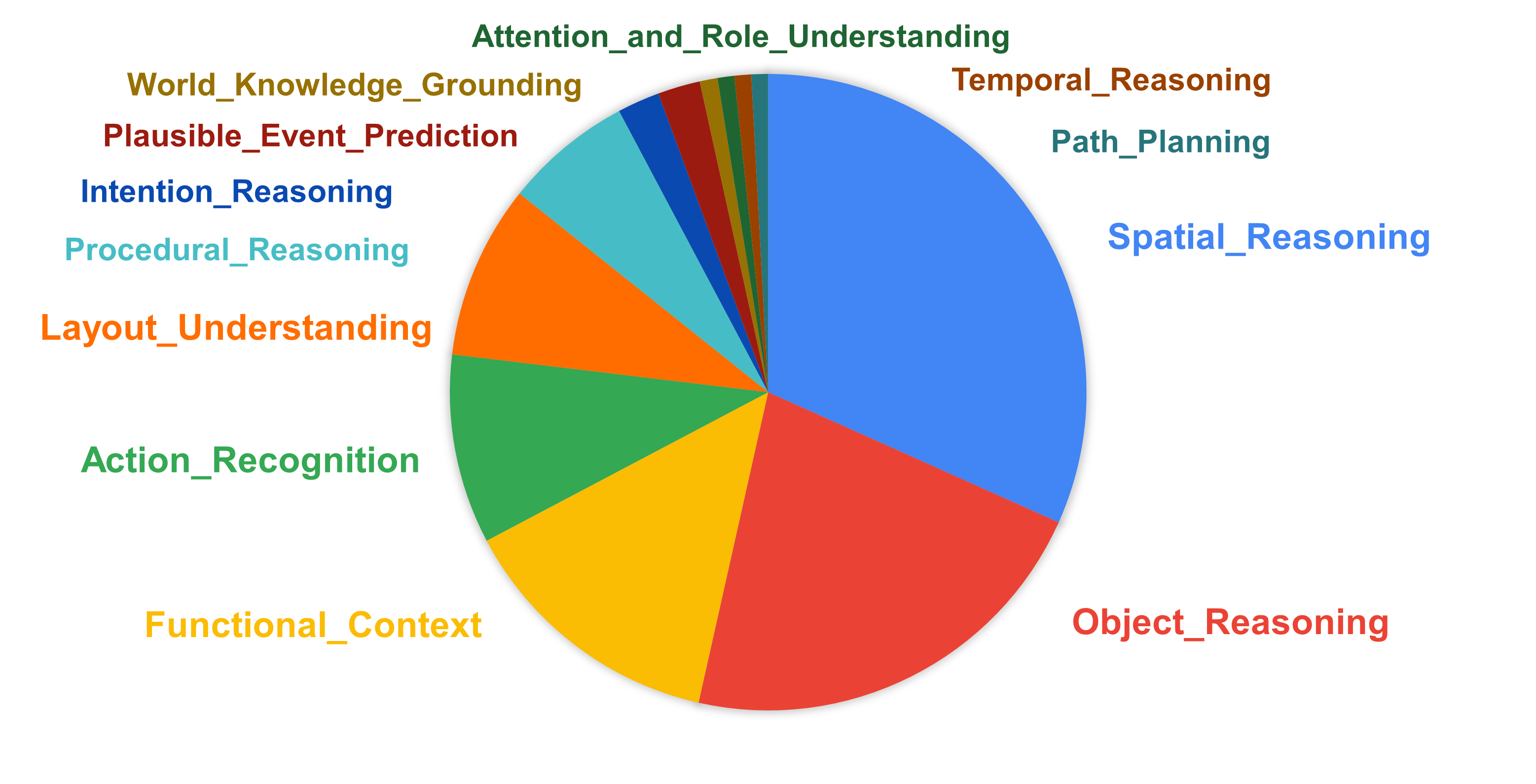}
    \caption{    
    Question reasoning type distribution over the dataset.
    }
    \label{fig:dataset_distribution}
\end{figure}

The dataset comprises a total of 1,766 QA instances with 630 images from the original ERQA~\cite{geminiroboticsteam2025geminiroboticsbringingai} dataset and 81 images curated from COIN~\cite{tang2019coinlargescaledatasetcomprehensive} and ScanNet~\cite{dai2017scannet}, spanning a broad spectrum of reasoning skills required by generalist embodied agents. The distribution across reasoning categories is shown in Figure~\ref{fig:dataset_distribution}.
% \input{table_statistics}

% Spatial Reasoning: 560 instances 
% Object Reasoning: 385 instances 
% Functional Context: 243 instances 
% Action Recognition: 170 instances 
% Layout Understanding: 156 instances 
% Procedural Reasoning: 115 instances 
% Intention Reasoning: 38 instances 
% Plausible Event Prediction: 38 instances 
% World Knowledge Grounding: 16 instances 
% Attention and Role Understanding: 15 instances
% Temporal Reasoning: 15 instances 
% Path Planning: 15 instances.

Overall, the dataset is dominated by spatial and object-centric reasoning, which together account for more than half of the benchmark. Higher-order cognitive skills—such as intention inference, event prediction, temporal reasoning, and path planning—are less frequent but deliberately included to evaluate advanced embodied reasoning capabilities beyond static perception. This distribution reflects a design choice to prioritize grounded visual understanding while ensuring coverage of planning, interaction, and anticipatory reasoning.

% \subsection{Human assessment}
% We design a structured human annotation interface to ensure consistent evaluation of QA samples. Annotators are presented with the image (visual context), the associated question-answer pair with multiple choices, the ground-truth label, and a brief explanation. A standardized 0–10 scoring rubric guides assessment, where higher scores indicate well-grounded, unambiguous, and logically coherent samples, and lower scores reflect ambiguity or errors. Annotators assign a score via a continuous scale and may provide optional comments.

\section{Experiments}
\paragraph{Experimental setup.} We adopt GPT-4o~\cite{openai2024gpt4ocard} as the default model for both QA generation and revision, and Gemini 3 Pro~\cite{google2025gemini3pro} as the default judge.
We conduct experiments on the ERQA-Plus dataset to evaluate a range of representative vision-language and embodied models under two settings: Multiple-Choice Question (MCQ) answering and open-ended question answering. We focus primarily on models in the $\sim$8B parameter regime, which we consider a practical ``Goldilocks zone'' for embodied AI. Models at this scale exhibit sufficiently strong semantic reasoning and instruction-following capabilities, while remaining computationally efficient enough for near real-time inference on edge hardware typically available in robotic systems. This trade-off ensures that our evaluation reflects realistic deployment constraints rather than purely theoretical performance. All experiments are conducted using NVIDIA A100 GPUs (80GB VRAM) and A40 GPUs (48GB VRAM).

\paragraph{VLM evaluation settings.}
We evaluate most vision-language models using the LlamaFactory framework~\cite{zheng2024llamafactory}. The evaluated models include LLaVA-NeXT-8B~\cite{liu2024llavanext}, MiniCPM-V-4.5-8B~\cite{yu2025minicpmv45cookingefficient}, Qwen3-VL-8B~\cite{qwen3technicalreport}. We also include Qwen3-VL-32B-Thinking to benchmark our 8B selections against a larger-scale reasoning model. Notably, we evaluate Prismatic-7B~\cite{karamcheti2024prismatic} using its official implementation to maintain consistency with its specialized inference pipeline. Prismatic-7B serves as a critical baseline in this study, as it functions as the architectural backbone for prominent Vision-Language-Action (VLA) models such as OpenVLA~\cite{kim24openvla} and OpenVLA-OFT~\cite{kim2025finetuningvisionlanguageactionmodelsoptimizing}, which bridge the gap between general-purpose vision-language understanding and embodied control. 

\paragraph{Embodied model evaluation settings.}
We evaluate RoboRefer-8B~\cite{zhou2025roborefer} and RoboBrain2.5-8B~\cite{tan2026robobrain25depthsight} following their official evaluation protocols. For LEO~\cite{huang2024embodied}, direct 3D point cloud inputs are not available in our dataset. To address this, we construct approximate 3D representations via a multi-stage pipeline: (1) depth estimation from RGB images using DepthAnything v3~\cite{lin2025depth3recoveringvisual}, (2) object proposal generation using GroundingDINO~\cite{liu2023grounding}, and (3) segmentation using SAM2~\cite{ravi2024sam2segmentimages}. These components are combined to produce object-centric point clouds, which are then used as input to LEO. We note that LEO processes only a single image per inference step, which may limit its ability to leverage multi-view contextual information.

% \paragraph{Evaluation metrics.}
% We evaluate model performance using two complementary metrics tailored to the task format.
% \textbf{MCQ accuracy.} For multiple-choice questions, we report standard accuracy, defined as the proportion of instances in which the model selects the correct option. Models are instructed to output only the option letter, which is extracted using simple regular expression matching.
% \textbf{Open-ended semantic similarity.} For open-ended responses, we adopt a semantic similarity metric based on Sentence-BERT (SBERT)~\cite{reimers-2019-sentence-bert}. Specifically, we encode both the model prediction and the ground-truth answer into embedding vectors using a pretrained SBERT model, and compute their cosine similarity. This metric captures semantic alignment beyond exact string matching and is robust to paraphrasing. Formally, given prediction embedding $\mathbf{p}$ and ground-truth embedding $\mathbf{g}$, the score is computed as:
% $    \text{Score} = \frac{\mathbf{p} \cdot \mathbf{g}}{\|\mathbf{p}\| \|\mathbf{g}\|}$.
% We report the mean cosine similarity across all samples. In this setting, models are instructed to provide direct answers without explanation, with responses constrained to a maximum of 20 words.
\paragraph{Evaluation metrics.}
We evaluate model performance using two complementary metrics tailored to the task format.
\textbf{MCQ accuracy.} For multiple-choice questions, we report standard accuracy (i.e., the proportion of correctly selected options). Models are instructed to output only the option letter, extracted via regular expression matching.
\textbf{Open-ended semantic similarity.} For open-ended responses, models are instructed to provide direct answers without explanation, constrained to at most 20 words. We use cosine similarity over Sentence-BERT (SBERT)~\cite{reimers-2019-sentence-bert} embeddings. Given prediction embedding $\mathbf{p}$ and ground-truth embedding $\mathbf{g}$, the score is:
$    \text{Score} = \frac{\mathbf{p} \cdot \mathbf{g}}{\|\mathbf{p}\| \|\mathbf{g}\|}$.
This metric captures semantic alignment beyond exact string matching and is robust to paraphrasing. We report the mean cosine similarity across all samples.

\begin{table}[htbp]
\centering
\caption{Model performance comparison: MCQ and open-ended evaluation. LEO's performance is showcased alongside other models; however, LEO results are greyed out to reflect a non-direct comparison due to varying experimental conditions.}
\label{tab:model_comparison_main}
\resizebox{\textwidth}{!}{%
\begin{tabular}{l c c c c c c c c c c c c c}
\toprule
\textbf{Model} & \textbf{Overall} & \textbf{Spatial} & \textbf{Object} & \textbf{Func.} & \textbf{Action} & \textbf{Layout} & \textbf{Proc.} & \textbf{Intent} & \textbf{Event} & \textbf{World} & \textbf{Role} & \textbf{Temp.} & \textbf{Path} \\
 & & \textbf{Reas.} & \textbf{Reas.} & \textbf{Cont.} & \textbf{Recog.} & \textbf{Under.} & \textbf{Reas.} & \textbf{Reas.} & \textbf{Pred.} & \textbf{Ground.} & \textbf{Under.} & \textbf{Reas.} & \textbf{Plan} \\
\midrule
\multicolumn{14}{l}{\textit{MCQ Evaluation}} \\
\midrule
LLaVA-NeXT-8B & 54.1 & 40.9 & 63.4 & 60.5 & 56.5 & 68.0 & 48.7 & 65.8 & 47.4 & 56.3 & 80.0 & 40.0 & 46.7 \\
Prismatic-7B & 64.6 & 53.8 & 75.8 & 74.9 & 57.1 & 72.4 & 61.7 & 71.1 & 68.4 & 62.5 & 86.7 & 33.3 & 26.7 \\
% MiniCPM-V-2.6-8B & 72.8 & 65.7 & 76.6 & 79.8 & 75.3 & 76.3 & 73.0 & 79.0 & 76.3 & 68.8 & 86.7 & 60.0 & 33.3 \\
MiniCPM-V-4.5-8B & 79.4 & 74.5 & 87.8 & 83.1 & 78.2 & 79.5 & 78.3 & 79.0 & 76.3 & 62.5 & 86.7 & 60.0 & 46.7 \\
Qwen3-VL-8B & 81.9 & 78.8 & 89.4 & 85.2 & 81.2 & 77.6 & 74.8 & 84.2 & 76.3 & 81.3 & 100.0 & 53.3 & 80.0 \\
\rowcolor{gray!20} Qwen3-VL-32B-Thinking & 83.4 & 80.7 & 84.9 & 88.5 & 85.9 & 82.1 & 80.0 & 92.1 & 86.8 & 93.8 & 93.3 & 46.7 & 60.0 \\
\hdashline
RoboRefer-8B & 69.6 & 65.0 & 75.6 & 75.3 & 67.1 & 76.3 & 60.9 & 68.4 & 68.4 & 50.0 & 73.3 & 53.3 & 60.0 \\
RoboBrain2.5-8B & 80.8 & 76.8 & 89.4 & 81.5 & 80.6 & 77.6 & 76.5 & 79.0 & 76.3 & 75.0 & 93.3 & 73.3 & 86.7 \\
% LEO & 4.1 & 4.1 & 4.2 & 4.1 & 4.7 & 1.9 & 7.8 & 0.0 & 0.0 & 6.3 & 6.7 & 0.0 & 13.3 \\
\color{gray} LEO & \color{gray} 4.1 & \color{gray} 4.1 & \color{gray} 4.2 & \color{gray} 4.1 & \color{gray} 4.7 & \color{gray} 1.9 & \color{gray} 7.8 & \color{gray} 0.0 & \color{gray} 0.0 & \color{gray} 6.3 & \color{gray} 6.7 & \color{gray} 0.0 & \color{gray} 13.3 \\
\midrule
\multicolumn{14}{l}{\textit{Open-Ended Evaluation}} \\
\midrule
LLaVA-NeXT-8B & 46.0 & 46.3 & 50.6 & 43.1 & 38.2 & 46.7 & 46.4 & 44.0 & 50.4 & 46.7 & 33.9 & 51.5 & 45.0 \\
Prismatic-7B & 51.8 & 53.9 & 62.1 & 47.0 & 43.2 & 49.4 & 44.3 & 43.1 & 44.3 & 44.5 & 42.3 & 46.6 & 33.3 \\
% MiniCPM-V-2.6-8B & 59.4 & 65.2 & 63.5 & 54.1 & 48.9 & 55.7 & 54.2 & 55.3 & 56.7 & 52.1 & 45.5 & 58.9 & 53.9 \\
MiniCPM-V-4.5-8B & 56.6 & 55.0 & 62.2 & 55.4 & 49.4 & 56.9 & 58.0 & 53.3 & 62.7 & 53.3 & 46.1 & 66.7 & 56.2 \\
Qwen3-VL-8B & 56.2 & 57.1 & 61.0 & 54.8 & 49.4 & 55.7 & 52.9 & 50.6 & 57.7 & 47.5 & 42.6 & 65.6 & 49.6 \\
\rowcolor{gray!20} Qwen3-VL-32B-Thinking & 61.4 & 65.1 & 67.7 & 55.0 & 53.6 & 56.0 & 60.9 & 54.4 & 58.4 & 55.4 & 50.4 & 66.1 & 55.6 \\
\hdashline
RoboRefer-8B & 47.7 & 48.2 & 53.0 & 45.7 & 39.1 & 48.1 & 44.7 & 40.9 & 54.3 & 47.1 & 36.4 & 49.8 & 44.7 \\
RoboBrain2.5-8B & 55.7 & 55.1 & 64.2 & 50.9 & 46.9 & 54.1 & 54.2 & 54.9 & 59.7 & 50.2 & 42.5 & 70.3 & 57.0 \\
\color{gray} LEO & \color{gray} 28.9 & \color{gray} 26.5 & \color{gray} 38.1 & \color{gray} 25.9 & \color{gray} 25.9 & \color{gray} 29.5 & \color{gray} 26.3 & \color{gray} 25.6 & \color{gray} 27.6 & \color{gray} 27.0 & \color{gray} 10.3 & \color{gray} 28.9 & \color{gray} 12.6 \\
\bottomrule
\end{tabular}%
}
\end{table}

\begin{table}[h!]
\centering
\caption{Model performance comparison across high-level reasoning categories.
\small \textbf{Abbreviations:} PR: Perceptual Reasoning, ACR: Action Centric Reasoning, CCR: Contextual and Commonsense Reasoning, NER: Navigation and Environmental Reasoning, SIR: Social and Interaction Reasoning.
}
\label{tab:model_performance_5types}
\resizebox{\textwidth}{!}{%
\begin{tabular}{l|ccccc|ccccc}
\toprule
\textbf{Model} 
& \multicolumn{5}{c|}{\textbf{MCQ}} 
& \multicolumn{5}{c}{\textbf{Open Ended}} \\
\cmidrule(lr){2-6} \cmidrule(lr){7-11}
& \textbf{PR} & \textbf{ACR} & \textbf{CCR} & \textbf{NER} & \textbf{SIR} 
& \textbf{PR} & \textbf{ACR} & \textbf{CCR} & \textbf{NER} & \textbf{SIR} \\
& (960) & (323) & (297) & (171) & (15) 
& (960) & (323) & (297) & (171) & (15) \\
\midrule
LLaVA-NeXT-8B & 50.0 & 54.8 & 58.6 & 66.1 & 80.0 & 48.1 & 41.8 & 44.2 & 46.6 & 33.9 \\
Prismatic-7B & 62.3 & 60.4 & 73.4 & 68.4 & 86.7 & 57.1 & 43.6 & 46.5 & 48.0 & 42.3 \\
% MiniCPM-V-2.6-8B & 70.0 & 74.9 & 78.8 & 72.5 & 86.7 & 64.4 & 51.6 & 54.4 & 55.5 & 45.5 \\
MiniCPM-V-4.5-8B & 79.5 & 78.3 & 81.1 & 76.6 & 86.7 & 58.1 & 52.9 & 56.2 & 56.8 & 46.1 \\
Qwen3-VL-8B & 82.5 & 79.3 & 83.8 & 77.8 & 100.0 & 58.8 & 50.8 & 54.8 & 55.2 & 42.6 \\
Qwen3-VL-32B-Thinking & 81.9 & 84.5 & 88.6 & 80.1 & 93.3 & 66.1 & 56.3 & 55.4 & 55.9 & 50.4 \\
\hdashline
RoboRefer-8B & 69.0 & 65.0 & 73.1 & 74.9 & 73.3 & 50.2 & 41.3 & 46.8 & 47.8 & 36.4 \\
RoboBrain2.5-8B & 81.7 & 79.0 & 80.5 & 78.4 & 93.3 & 59.0 & 50.4 & 52.0 & 54.3 & 42.5 \\
\color{gray} LEO & \color{gray} 4.1 & \color{gray} 5.3 & \color{gray} 3.7 & \color{gray} 2.9 & \color{gray} 6.7 & \color{gray} 31.2 & \color{gray} 26.0 & \color{gray} 26.2 & \color{gray} 28.0 & \color{gray} 10.3 \\
\bottomrule
\end{tabular}%
}
\vspace{1em}

\end{table}

\subsection{Experimental results}
Tables \ref{tab:model_comparison_main} and \ref{tab:model_performance_5types} present model performance across 12 fine-grained reasoning subtypes and 5 high-level reasoning categories, under both MCQ and open-ended evaluation (measured by SBERT semantic scores) on a benchmark of 1,766 questions. Qualitative results are shown in the Appendix.

\paragraph{Performance hierarchy and evaluation consistency} Across all questions spanning reasoning subtypes under both MCQ and open-ended evaluation, a clear performance hierarchy emerges. Qwen3-VL-32B-Thinking leads overall at 83.4\% MCQ and 61.4\% open-ended, benefiting from scale and explicit chain-of-thought reasoning. The strong concordance between MCQ accuracy and open-ended SBERT scores across all models confirms that benchmark signals are consistent across evaluation protocols. However, inference speed of Qwen3-VL-32B-Thinking is poor.

\paragraph{The 8B-scale frontier} Among $\sim$8B models, Qwen3-VL-8B and MiniCPM-V-4.5-8B establish a strong general-VLM baseline (>79\% MCQ, $\sim$56\% open-ended). RoboBrain2.5-8B matches this cohort closely (80.8\% MCQ, 55.7\% open-ended), demonstrating that robotics-oriented training does not sacrifice breadth. RoboRefer-8B, however, lags significantly (69.6\% MCQ, 47.7\% open-ended), cautioning that domain-specific training does not uniformly confer generalisation benefits.

\paragraph{Scale does not resolve hard subtypes} Temporal Reasoning and Path Planning emerge as the benchmark's most discriminative subtypes, with several models scoring below chance on MCQ. Critically, even Qwen3-VL-32B-Thinking drops to 46.7\% on Temporal Reasoning—its worst subtype—confirming that multi-image temporal understanding remains unsolved by scale alone.

\paragraph{Robotics-specific training as efficient inductive bias} RoboBrain2.5-8B's performance on these hard subtypes is the study's most significant finding. It achieves 73.3\% on Temporal Reasoning MCQ—exceeding Qwen3-VL-32B-Thinking by +26.6 points and the next-best 8B model by +20.0—and leads open-ended evaluation on the same subtype (70.3 vs. 66.1). On Path Planning, it similarly leads all models at 86.7\% MCQ. These margins demonstrate that domain-aligned training can compensate for substantial capacity differences, offering a compelling efficiency argument for specialised architectures.

\paragraph{Cross-modal asymmetries and remaining bottlenecks} A recurring inversion across modalities reveals important distinctions between recognition and generation competencies: Social and Interaction Reasoning ranks highly under MCQ yet becomes the hardest open-ended category (best: 50.4\%), while Perceptual Reasoning shows the inverse pattern. Action-Centric Reasoning remains a shared bottleneck across all architectures in open-ended generation (<52\% for most 8B models), pointing to procedural and intentional reasoning as a key target for future embodied VQA research.

\subsection{Ablation on QA creation}
\begin{figure}[htbp]
     \centering
     \begin{subfigure}[b]{0.49\textwidth}
         \centering
         \includegraphics[width=\textwidth]{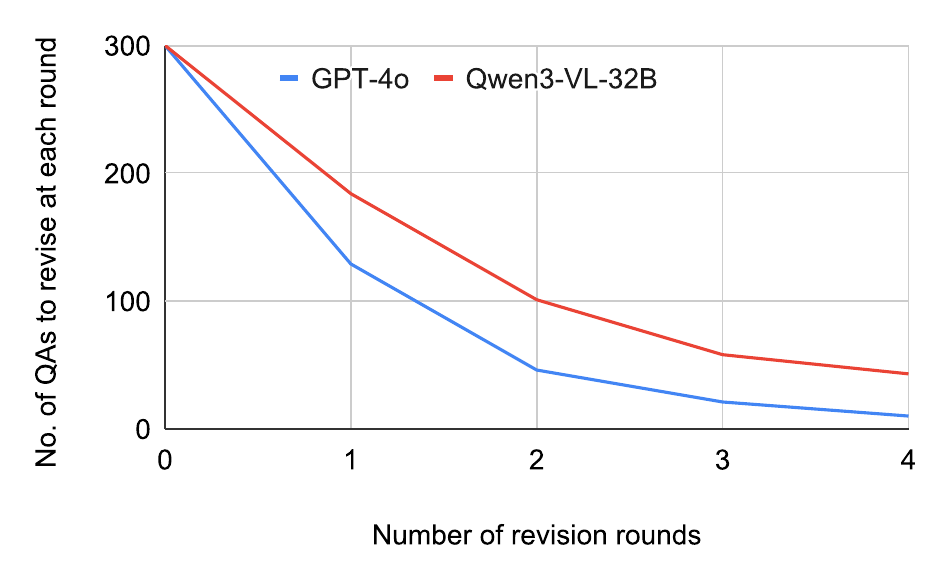}
         \label{fig:ablation_left}
     \end{subfigure}
     % \hfill
     \begin{subfigure}[b]{0.49\textwidth}
         \centering
         % \includegraphics[width=\textwidth]{figures/average_score_at_each_round.png}
         %\includesvg[width=\textwidth]{figures/ablation_right}
         \includegraphics[width=\textwidth]{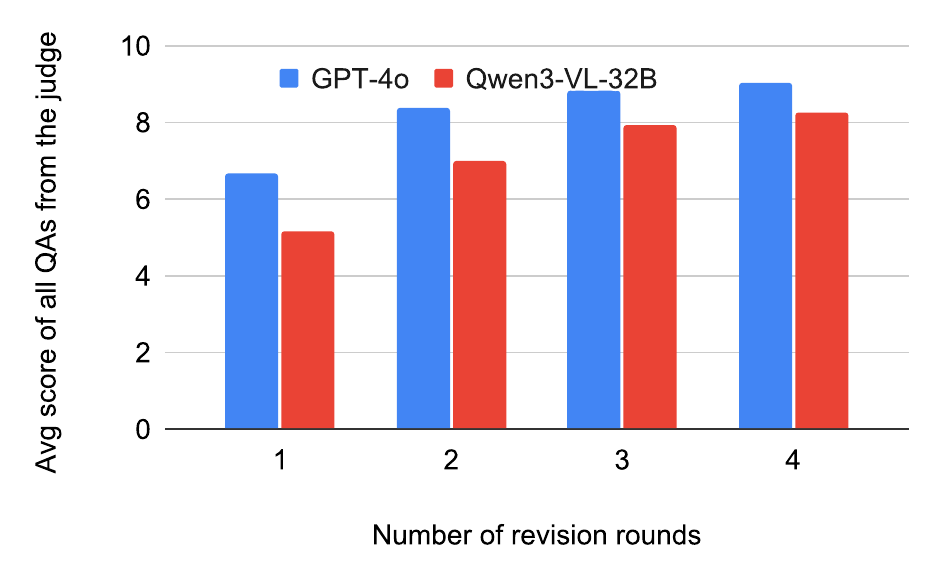}
         \label{fig:ablation_right}
     \end{subfigure}
     \caption{Ablation study on iterative revision across rounds. \textbf{Left:} the number of QA pairs requiring revision decreases over rounds for both GPT-4o and Qwen3-VL-32B, demonstrating convergence. \textbf{Right:} judge scores improve progressively for both models, with GPT-4o consistently achieving higher quality across all rounds. We use Gemini 3 Pro as the judge for both models.}
     \label{fig:ablation_figure}
\end{figure}

We curate a balanced subset of 300 QA pairs, uniformly sampled across all reasoning categories, to systematically analyze the contributions of each component in our pipeline. We conduct two complementary ablation studies focusing on the generator/reviser and judge modules.
Figure~\ref{fig:ablation_figure} illustrates the dynamics of iterative revision. Across revision rounds, the number of QA pairs requiring further refinement decreases monotonically for both models, indicating stable convergence. At the same time, the average quality scores assigned by the judge improve consistently, validating the effectiveness of the iterative refinement process.

\paragraph{Generator comparison}
To evaluate the impact of the QA generator, we compare GPT-4o and Qwen3-VL-32B under a controlled setup where Qwen3-VL-32B is used as the reviser and Gemini 3 Pro serves as the judge.

As shown in Figure~\ref{fig:ablation_figure}, GPT-4o demonstrates faster convergence, with a steeper reduction in the number of QA pairs requiring revision. By the second round, GPT-4o reduces the revision set significantly more than Qwen3-VL-32B, indicating that it produces higher-quality initial QA pairs that require fewer corrections. In addition, GPT-4o consistently achieves higher judge scores across all rounds. This suggests that its outputs are not only easier to refine but also reach a higher final quality ceiling. In contrast, Qwen3-VL-32B exhibits slower convergence and lower final scores, implying weaker initial generation quality and less effective refinement trajectories. Overall, these results highlight that the choice of generator and reviser play a critical role in both efficiency (fewer revision steps) and final QA quality.

\paragraph{Judge comparison} We further assess the reliability of LLM-based judges by comparing their evaluations against human annotations on the same 300-sample subset. 
We design a structured human annotation interface to ensure consistent evaluation of QA samples. Annotators are presented with the image (visual context), the associated question-answer pair with multiple choices, the ground-truth label, and a brief explanation. A standardized 0–10 scoring rubric guides assessment, where higher scores indicate well-grounded, unambiguous, and logically coherent samples, and lower scores reflect ambiguity or errors. Annotators assign a score via a continuous scale and may provide optional comments.
Specifically, we evaluate Gemini 3 Pro and Qwen3-VL-32B as judges using correlation-based, exact-matching-based, and error-based metrics.

Gemini 3 Pro demonstrates stronger alignment with human judgment, achieving a higher Pearson correlation (0.42) and exact-matching accuracy (0.68). This indicates better sensitivity to relative quality differences and improved ranking capability.
In contrast, Qwen3-VL-32B shows weak correlation (0.18) and lower exact-matching accuracy (0.53), suggesting limited ability to distinguish high- and low-quality samples. Interestingly, it achieves a slightly lower mean absolute error (2.28 vs. 2.38), which can be attributed to its conservative scoring behavior. Its higher mean and reduced variance indicate a tendency to compress predictions toward the center, leading to lower absolute error but poor discriminative power.

These findings reveal a fundamental trade-off: Gemini 3 Pro provides better ranking fidelity but is less calibrated, whereas Qwen3-VL-32B is more stable but less informative. Importantly, neither model achieves strong agreement with human evaluators, suggesting that current LLM-based judges are insufficient as standalone evaluators and may require calibration or hybrid human-in-the-loop strategies.

\section{Related work}
\begin{comment}
2.1 Visual Question Answering Benchmarks\\
Programmatically generated datasets\\
Web-sourced / crowdsourced datasets\\
Manually curated datasets\\
Limitations: shortcut learning, answer bias, shallow reasoning\\
2.2 Embodied question answering (EQA)\\
Simulator-based EQA\\
Navigation + QA benchmarks\\
World-model-based embodied agents\\
Gaps: limited reasoning depth, weak dependency modeling\\    
\end{comment}

\paragraph{Early embodied question answering}
Embodied Question Answering (EQA) studies agents that perceive, navigate, and reason in environments to answer natural-language questions~\cite{ahn2022can}. Early work~\cite{das2018embodied} introduced EQA in synthetic 3D environments (e.g., House3D~\cite{wu2018building}), where agents gather egocentric observations to answer programmatically generated questions, enabling controlled distributions and reduced language bias. Neural Modular Control~\cite{das2018neural} further decomposed EQA into navigation and QA modules, emphasizing the integration of perception, action, and reasoning.

\paragraph{Toward realistic and open-ended EQA}
Recent work shifts from synthetic setups to more realistic and open-ended settings. OpenEQA~\cite{majumdar2024openeqa} introduces human-generated, open-vocabulary questions across real-world environments, supporting both memory-based and active exploration scenarios with LLM-based evaluation. This transition reflects a broader move toward scalable, foundation-model-driven embodied understanding.

\paragraph{Increasing reasoning complexity}
Several works extend EQA to require deeper reasoning. Knowledge-based EQA~\cite{tan2023knowledge} incorporates external knowledge and logical reasoning beyond visible scene content. 3D-MoRe~\cite{xu20253d} focuses on multimodal reasoning over 3D scenes by integrating language with structured 3D representations, highlighting the importance of geometric grounding.

\paragraph{Multimodal and interactive settings}
Other approaches emphasize richer interaction modalities. EQA-MX~\cite{islam2023eqa} introduces multimodal QA with verbal and gestural queries, while DialFRED~\cite{gao2022dialfred} embeds question answering within interactive, dialogue-driven embodied tasks, enabling clarification and human-agent collaboration.

\paragraph{Robustness and exploration-aware evaluation}
Recent benchmarks focus on robustness and evaluation quality. NoisyEQA~\cite{wu2024noisyeqa} studies performance under ambiguous or noisy queries, while EXPRESS-Bench~\cite{jiang2025beyond} evaluates whether exploration behavior is relevant and sufficient for answering questions, linking reasoning to evidence collection.

\paragraph{Affordance grounding and long-horizon reasoning}
SayCan~\cite{ahn2022can} grounds language in executable robot affordances to answer action-oriented queries, while RoboVQA~\cite{sermanet2024robovqa} leverages human-robot demonstrations for temporally grounded, multi-step reasoning.

\paragraph{Summary}
Overall, EQA has evolved from synthetic, closed-vocabulary settings to real-world, open-ended, and multimodal scenarios, with increasing emphasis on reasoning complexity, robustness, and interaction. However, most benchmarks still focus on answer correctness after exploration, with limited control over the specific reasoning processes being evaluated.

\section{Discussion \& conclusions}

\paragraph{Data limitations}
ERQA-Plus is a diagnostic benchmark rather than a full simulation of embodied intelligence. Its coverage is broad but uneven, with spatial and object-centric reasoning dominating, while higher-order skills (e.g., intention, temporal reasoning, path planning) remain underrepresented and should be interpreted cautiously. The dataset relies on static RGB images, evaluating visual inference but not closed-loop control, execution, or long-horizon interaction; approximated 3D inputs may further introduce noise for some models. The generator--judge--reviser pipeline improves scalability but may introduce biases, and automated judges show limited agreement with humans. The benchmark is also limited to English MCQ and short-answer formats, motivating future extensions to richer modalities, better category balance, and interactive, multilingual settings.

\paragraph{Usage}
ERQA-Plus is best used as a fine-grained diagnostic tool rather than a single-score leaderboard. Aggregate metrics (e.g., accuracy, SBERT) should be complemented with category-level analysis to reveal specific reasoning strengths and weaknesses. It supports evaluation of both VLMs and embodied models, targeted debugging via its taxonomy, and reuse of its generation pipeline for creating grounded QA datasets.

\paragraph{Social impact}
ERQA-Plus can improve safety and reliability by identifying reasoning failures before deployment in real-world environments. It also encourages more cautious interpretation of progress, showing that strong VQA or language performance does not imply grounded reasoning. However, it should not be used in isolation, as models may overfit; complementary evaluation (simulation, real-world testing, human assessment) is essential. Responsible use must also address privacy, consent, data governance, and potential biases in model-assisted data generation.

\paragraph{Conclusion}
ERQA-Plus introduces a structured, quality-controlled benchmark for evaluating embodied reasoning across 1,766 QA pairs and multiple reasoning categories. Results show that current models, while improving, remain inconsistent—especially in spatial, procedural, and predictive reasoning—highlighting the limits of standard VQA evaluation. By enabling fine-grained analysis and exposing failure modes, ERQA-Plus provides a foundation for more interpretable evaluation and the development of robust embodied agents, with future work focusing on richer modalities, broader coverage, and interactive settings.

\noindent{\textbf{Acknowledgements.} This research/project is supported by the National Research Foundation, Singapore, under its NRF Fellowship (Award\# NRF-NRFF14-2022-0001). This research is also supported by funding allocation to B.F. by the Agency for Science, Technology and Research (A*STAR) under its SERC Central Research Fund (CRF), as well as its Centre for Frontier AI Research (CFAR).}

\bibliographystyle{abbrv}
\bibliography{main}

@inproceedings{majumdar2024openeqa,
  title={Openeqa: Embodied question answering in the era of foundation models},
  author={Majumdar, Arjun and Ajay, Anurag and Zhang, Xiaohan and Putta, Pranav and Yenamandra, Sriram and Henaff, Mikael and Silwal, Sneha and Mcvay, Paul and Maksymets, Oleksandr and Arnaud, Sergio and others},
  booktitle={Proceedings of the IEEE/CVF conference on computer vision and pattern recognition},
  pages={16488--16498},
  year={2024}
}

@inproceedings{das2018embodied,
  title={Embodied question answering},
  author={Das, Abhishek and Datta, Samyak and Gkioxari, Georgia and Lee, Stefan and Parikh, Devi and Batra, Dhruv},
  booktitle={Proceedings of the IEEE conference on computer vision and pattern recognition},
  pages={1--10},
  year={2018}
}

@article{tan2023knowledge,
  title={Knowledge-based embodied question answering},
  author={Tan, Sinan and Ge, Mengmeng and Guo, Di and Liu, Huaping and Sun, Fuchun},
  journal={IEEE Transactions on Pattern Analysis and Machine Intelligence},
  volume={45},
  number={10},
  pages={11948--11960},
  year={2023},
  publisher={IEEE}
}

@inproceedings{islam2023eqa,
  title={Eqa-mx: Embodied question answering using multimodal expression},
  author={Islam, Md Mofijul and Gladstone, Alexi and Islam, Riashat and Iqbal, Tariq},
  booktitle={The Twelfth International Conference on Learning Representations},
  year={2023}
}

@inproceedings{jiang2025beyond,
  title={Beyond the destination: A novel benchmark for exploration-aware embodied question answering},
  author={Jiang, Kaixuan and Liu, Yang and Chen, Weixing and Luo, Jingzhou and Chen, Ziliang and Pan, Ling and Li, Guanbin and Lin, Liang},
  booktitle={Proceedings of the IEEE/CVF International Conference on Computer Vision},
  pages={9091--9101},
  year={2025}
}

@inproceedings{xu20253d,
  title={3d-more: Unified modal-contextual reasoning for embodied question answering},
  author={Xu, Rongtao and Gao, Han and Yu, Mingming and An, Dong and Chen, Shunpeng and Wang, Changwei and Guo, Li and Liang, Xiaodan and Xu, Shibiao},
  booktitle={2025 IEEE/RSJ International Conference on Intelligent Robots and Systems (IROS)},
  pages={5924--5929},
  year={2025},
  organization={IEEE}
}

@article{wu2024noisyeqa,
  title={Noisyeqa: Benchmarking embodied question answering against noisy queries},
  author={Wu, Tao and Zhou, Chuhao and Wong, Yen Heng and Gu, Lin and Yang, Jianfei},
  journal={arXiv preprint arXiv:2412.10726},
  year={2024}
}

@article{gao2022dialfred,
  title={Dialfred: Dialogue-enabled agents for embodied instruction following},
  author={Gao, Xiaofeng and Gao, Qiaozi and Gong, Ran and Lin, Kaixiang and Thattai, Govind and Sukhatme, Gaurav S},
  journal={IEEE Robotics and Automation Letters},
  volume={7},
  number={4},
  pages={10049--10056},
  year={2022},
  publisher={IEEE}
}

@inproceedings{das2018neural,
  title={Neural modular control for embodied question answering},
  author={Das, Abhishek and Gkioxari, Georgia and Lee, Stefan and Parikh, Devi and Batra, Dhruv},
  booktitle={Conference on robot learning},
  pages={53--62},
  year={2018},
  organization={PMLR}
}

@misc{tang2019coinlargescaledatasetcomprehensive,
      title={COIN: A Large-scale Dataset for Comprehensive Instructional Video Analysis}, 
      author={Yansong Tang and Dajun Ding and Yongming Rao and Yu Zheng and Danyang Zhang and Lili Zhao and Jiwen Lu and Jie Zhou},
      year={2019},
      eprint={1903.02874},
      archivePrefix={arXiv},
      primaryClass={cs.CV},
      url={https://arxiv.org/abs/1903.02874}, 
}

@inproceedings{dai2017scannet,
    title={ScanNet: Richly-annotated 3D Reconstructions of Indoor Scenes},
    author={Dai, Angela and Chang, Angel X. and Savva, Manolis and Halber, Maciej and Funkhouser, Thomas and Nie{\ss}ner, Matthias},
    booktitle = {Proc. Computer Vision and Pattern Recognition (CVPR), IEEE},
    year = {2017}
}

@misc{geminiroboticsteam2025geminiroboticsbringingai,
      title={Gemini Robotics: Bringing AI into the Physical World}, 
      author={Gemini Robotics Team and Saminda Abeyruwan and Joshua Ainslie and Jean-Baptiste Alayrac and Montserrat Gonzalez Arenas and Travis Armstrong and Ashwin Balakrishna and Robert Baruch and Maria Bauza and Michiel Blokzijl and Steven Bohez and Konstantinos Bousmalis and Anthony Brohan and Thomas Buschmann and Arunkumar Byravan and Serkan Cabi and Ken Caluwaerts and Federico Casarini and Oscar Chang and Jose Enrique Chen and Xi Chen and Hao-Tien Lewis Chiang and Krzysztof Choromanski and David D'Ambrosio and Sudeep Dasari and Todor Davchev and Coline Devin and Norman Di Palo and Tianli Ding and Adil Dostmohamed and Danny Driess and Yilun Du and Debidatta Dwibedi and Michael Elabd and Claudio Fantacci and Cody Fong and Erik Frey and Chuyuan Fu and Marissa Giustina and Keerthana Gopalakrishnan and Laura Graesser and Leonard Hasenclever and Nicolas Heess and Brandon Hernaez and Alexander Herzog and R. Alex Hofer and Jan Humplik and Atil Iscen and Mithun George Jacob and Deepali Jain and Ryan Julian and Dmitry Kalashnikov and M. Emre Karagozler and Stefani Karp and Chase Kew and Jerad Kirkland and Sean Kirmani and Yuheng Kuang and Thomas Lampe and Antoine Laurens and Isabel Leal and Alex X. Lee and Tsang-Wei Edward Lee and Jacky Liang and Yixin Lin and Sharath Maddineni and Anirudha Majumdar and Assaf Hurwitz Michaely and Robert Moreno and Michael Neunert and Francesco Nori and Carolina Parada and Emilio Parisotto and Peter Pastor and Acorn Pooley and Kanishka Rao and Krista Reymann and Dorsa Sadigh and Stefano Saliceti and Pannag Sanketi and Pierre Sermanet and Dhruv Shah and Mohit Sharma and Kathryn Shea and Charles Shu and Vikas Sindhwani and Sumeet Singh and Radu Soricut and Jost Tobias Springenberg and Rachel Sterneck and Razvan Surdulescu and Jie Tan and Jonathan Tompson and Vincent Vanhoucke and Jake Varley and Grace Vesom and Giulia Vezzani and Oriol Vinyals and Ayzaan Wahid and Stefan Welker and Paul Wohlhart and Fei Xia and Ted Xiao and Annie Xie and Jinyu Xie and Peng Xu and Sichun Xu and Ying Xu and Zhuo Xu and Yuxiang Yang and Rui Yao and Sergey Yaroshenko and Wenhao Yu and Wentao Yuan and Jingwei Zhang and Tingnan Zhang and Allan Zhou and Yuxiang Zhou},
      year={2025},
      eprint={2503.20020},
      archivePrefix={arXiv},
      primaryClass={cs.RO},
      url={https://arxiv.org/abs/2503.20020}, 
}

@inproceedings{zheng2024llamafactory,
  title={LlamaFactory: Unified Efficient Fine-Tuning of 100+ Language Models},
  author={Yaowei Zheng and Richong Zhang and Junhao Zhang and Yanhan Ye and Zheyan Luo and Zhangchi Feng and Yongqiang Ma},
  booktitle={Proceedings of the 62nd Annual Meeting of the Association for Computational Linguistics (Volume 3: System Demonstrations)},
  address={Bangkok, Thailand},
  publisher={Association for Computational Linguistics},
  year={2024},
  url={http://arxiv.org/abs/2403.13372}
}

@misc{liu2024llavanext,
    title={LLaVA-NeXT: Improved reasoning, OCR, and world knowledge},
    url={https://llava-vl.github.io/blog/2024-01-30-llava-next/},
    author={Liu, Haotian and Li, Chunyuan and Li, Yuheng and Li, Bo and Zhang, Yuanhan and Shen, Sheng and Lee, Yong Jae},
    month={January},
    year={2024}
}

@misc{yu2025minicpmv45cookingefficient,
      title={MiniCPM-V 4.5: Cooking Efficient MLLMs via Architecture, Data, and Training Recipe}, 
      author={Tianyu Yu and Zefan Wang and Chongyi Wang and Fuwei Huang and Wenshuo Ma and Zhihui He and Tianchi Cai and Weize Chen and Yuxiang Huang and Yuanqian Zhao and Bokai Xu and Junbo Cui and Yingjing Xu and Liqing Ruan and Luoyuan Zhang and Hanyu Liu and Jingkun Tang and Hongyuan Liu and Qining Guo and Wenhao Hu and Bingxiang He and Jie Zhou and Jie Cai and Ji Qi and Zonghao Guo and Chi Chen and Guoyang Zeng and Yuxuan Li and Ganqu Cui and Ning Ding and Xu Han and Yuan Yao and Zhiyuan Liu and Maosong Sun},
      year={2025},
      eprint={2509.18154},
      archivePrefix={arXiv},
      primaryClass={cs.LG},
      url={https://arxiv.org/abs/2509.18154}, 
}

@misc{qwen3technicalreport,
      title={Qwen3 Technical Report}, 
      author={Qwen Team},
      year={2025},
      eprint={2505.09388},
      archivePrefix={arXiv},
      primaryClass={cs.CL},
      url={https://arxiv.org/abs/2505.09388}, 
}

@inproceedings{karamcheti2024prismatic,
  title = {Prismatic VLMs: Investigating the Design Space of Visually-Conditioned Language Models},
  author = {Siddharth Karamcheti and Suraj Nair and Ashwin Balakrishna and Percy Liang and Thomas Kollar and Dorsa Sadigh},
  booktitle = {International Conference on Machine Learning (ICML)},
  year = {2024},
}

@article{zhou2025roborefer,
    title={RoboRefer: Towards Spatial Referring with Reasoning in Vision-Language Models for Robotics},
    author={Zhou, Enshen and An, Jingkun and Chi, Cheng and Han, Yi and Rong, Shanyu and Zhang, Chi and Wang, Pengwei and Wang, Zhongyuan and Huang, Tiejun and Sheng, Lu and others},
    journal={arXiv preprint arXiv:2506.04308},
    year={2025}
}

@article{tan2026robobrain25depthsight,
  title={RoboBrain 2.5: Depth in Sight, Time in Mind},
  author={Tan, Huajie and Zhou, Enshen and Li, Zhiyu and Xu, Yijie and Ji, Yuheng and Chen, Xiansheng and Chi, Cheng and Wang, Pengwei and Jia, Huizhu and Ao, Yulong and Cao, Mingyu and Chen, Sixiang and Li, Zhe and Liu, Mengzhen and Wang, Zixiao and Rong, Shanyu and Lyu, Yaoxu and Zhao, Zhongxia and Co, Peterson and Li, Yibo and Han, Yi and Xie, Shaoxuan and Yao, Guocai and Wang, Songjing and Zhang, Leiduo and Yang, Xi and Jiao, Yance and Shi, Donghai and Xie, Kunchang and Nie, Shaokai and Men, Chunlei and Lin, Yonghua and Wang, Zhongyuan and Huang, Tiejun and Zhang, Shanghang},
  journal={arXiv preprint arXiv:2601.14352},
  year={2026}
}

@inproceedings{huang2024embodied,
  title={An Embodied Generalist Agent in 3D World},
  author={Huang, Jiangyong and Yong, Silong and Ma, Xiaojian and Linghu, Xiongkun and Li, Puhao and Wang, Yan and Li, Qing and Zhu, Song-Chun and Jia, Baoxiong and Huang, Siyuan},
  booktitle={Proceedings of the International Conference on Machine Learning (ICML)},
  year={2024}
}

@misc{lin2025depth3recoveringvisual,
      title={Depth Anything 3: Recovering the Visual Space from Any Views}, 
      author={Haotong Lin and Sili Chen and Junhao Liew and Donny Y. Chen and Zhenyu Li and Guang Shi and Jiashi Feng and Bingyi Kang},
      year={2025},
      eprint={2511.10647},
      archivePrefix={arXiv},
      primaryClass={cs.CV},
      url={https://arxiv.org/abs/2511.10647}, 
}

@article{liu2023grounding,
  title={Grounding dino: Marrying dino with grounded pre-training for open-set object detection},
  author={Liu, Shilong and Zeng, Zhaoyang and Ren, Tianhe and Li, Feng and Zhang, Hao and Yang, Jie and Li, Chunyuan and Yang, Jianwei and Su, Hang and Zhu, Jun and others},
  journal={arXiv preprint arXiv:2303.05499},
  year={2023}
}

@misc{ravi2024sam2segmentimages,
      title={SAM 2: Segment Anything in Images and Videos}, 
      author={Nikhila Ravi and Valentin Gabeur and Yuan-Ting Hu and Ronghang Hu and Chaitanya Ryali and Tengyu Ma and Haitham Khedr and Roman Rädle and Chloe Rolland and Laura Gustafson and Eric Mintun and Junting Pan and Kalyan Vasudev Alwala and Nicolas Carion and Chao-Yuan Wu and Ross Girshick and Piotr Dollár and Christoph Feichtenhofer},
      year={2024},
      eprint={2408.00714},
      archivePrefix={arXiv},
      primaryClass={cs.CV},
      url={https://arxiv.org/abs/2408.00714}, 
}

@article{ahn2022can,
  title={Do as i can, not as i say: Grounding language in robotic affordances},
  author={Ahn, Michael and Brohan, Anthony and Brown, Noah and Chebotar, Yevgen and Cortes, Omar and David, Byron and Finn, Chelsea and Fu, Chuyuan and Gopalakrishnan, Keerthana and Hausman, Karol and others},
  journal={arXiv preprint arXiv:2204.01691},
  year={2022}
}

@inproceedings{sermanet2024robovqa,
  title={Robovqa: Multimodal long-horizon reasoning for robotics},
  author={Sermanet, Pierre and Ding, Tianli and Zhao, Jeffrey and Xia, Fei and Dwibedi, Debidatta and Gopalakrishnan, Keerthana and Chan, Christine and Dulac-Arnold, Gabriel and Maddineni, Sharath and Joshi, Nikhil J and others},
  booktitle={2024 IEEE International Conference on Robotics and Automation (ICRA)},
  pages={645--652},
  year={2024},
  organization={IEEE}
}

@inproceedings{reimers-2019-sentence-bert,
    title = "Sentence-BERT: Sentence Embeddings using Siamese BERT-Networks",
    author = "Reimers, Nils and Gurevych, Iryna",
    booktitle = "Proceedings of the 2019 Conference on Empirical Methods in Natural Language Processing",
    month = "11",
    year = "2019",
    publisher = "Association for Computational Linguistics",
    url = "https://arxiv.org/abs/1908.10084",
}

@article{wu2018building,
  title={Building generalizable agents with a realistic and rich 3D environment},
  author={Wu, Yi and Wu, Yuxin and Gkioxari, Georgia and Tian, Yuandong},
  journal={arXiv preprint arXiv:1801.02209},
  year={2018}
}

@article{kim24openvla,
    title={OpenVLA: An Open-Source Vision-Language-Action Model},
    author={{Moo Jin} Kim and Karl Pertsch and Siddharth Karamcheti and Ted Xiao and Ashwin Balakrishna and Suraj Nair and Rafael Rafailov and Ethan Foster and Grace Lam and Pannag Sanketi and Quan Vuong and Thomas Kollar and Benjamin Burchfiel and Russ Tedrake and Dorsa Sadigh and Sergey Levine and Percy Liang and Chelsea Finn},
    journal = {arXiv preprint arXiv:2406.09246},
    year={2024}
}

@misc{kim2025finetuningvisionlanguageactionmodelsoptimizing,
      title={Fine-Tuning Vision-Language-Action Models: Optimizing Speed and Success}, 
      author={Moo Jin Kim and Chelsea Finn and Percy Liang},
      year={2025},
      eprint={2502.19645},
      archivePrefix={arXiv},
      primaryClass={cs.RO},
      url={https://arxiv.org/abs/2502.19645}, 
}

@misc{openai2024gpt4ocard,
      title={GPT-4o System Card}, 
      author={OpenAI and : and Aaron Hurst and Adam Lerer and Adam P. Goucher and Adam Perelman and Aditya Ramesh and Aidan Clark and AJ Ostrow and Akila Welihinda and Alan Hayes and Alec Radford and Aleksander Mądry and Alex Baker-Whitcomb and Alex Beutel and Alex Borzunov and Alex Carney and Alex Chow and Alex Kirillov and Alex Nichol and Alex Paino and Alex Renzin and Alex Tachard Passos and Alexander Kirillov and Alexi Christakis and Alexis Conneau and Ali Kamali and Allan Jabri and Allison Moyer and Allison Tam and Amadou Crookes and Amin Tootoochian and Amin Tootoonchian and Ananya Kumar and Andrea Vallone and Andrej Karpathy and Andrew Braunstein and Andrew Cann and Andrew Codispoti and Andrew Galu and Andrew Kondrich and Andrew Tulloch and Andrey Mishchenko and Angela Baek and Angela Jiang and Antoine Pelisse and Antonia Woodford and Anuj Gosalia and Arka Dhar and Ashley Pantuliano and Avi Nayak and Avital Oliver and Barret Zoph and Behrooz Ghorbani and Ben Leimberger and Ben Rossen and Ben Sokolowsky and Ben Wang and Benjamin Zweig and Beth Hoover and Blake Samic and Bob McGrew and Bobby Spero and Bogo Giertler and Bowen Cheng and Brad Lightcap and Brandon Walkin and Brendan Quinn and Brian Guarraci and Brian Hsu and Bright Kellogg and Brydon Eastman and Camillo Lugaresi and Carroll Wainwright and Cary Bassin and Cary Hudson and Casey Chu and Chad Nelson and Chak Li and Chan Jun Shern and Channing Conger and Charlotte Barette and Chelsea Voss and Chen Ding and Cheng Lu and Chong Zhang and Chris Beaumont and Chris Hallacy and Chris Koch and Christian Gibson and Christina Kim and Christine Choi and Christine McLeavey and Christopher Hesse and Claudia Fischer and Clemens Winter and Coley Czarnecki and Colin Jarvis and Colin Wei and Constantin Koumouzelis and Dane Sherburn and Daniel Kappler and Daniel Levin and Daniel Levy and David Carr and David Farhi and David Mely and David Robinson and David Sasaki and Denny Jin and Dev Valladares and Dimitris Tsipras and Doug Li and Duc Phong Nguyen and Duncan Findlay and Edede Oiwoh and Edmund Wong and Ehsan Asdar and Elizabeth Proehl and Elizabeth Yang and Eric Antonow and Eric Kramer and Eric Peterson and Eric Sigler and Eric Wallace and Eugene Brevdo and Evan Mays and Farzad Khorasani and Felipe Petroski Such and Filippo Raso and Francis Zhang and Fred von Lohmann and Freddie Sulit and Gabriel Goh and Gene Oden and Geoff Salmon and Giulio Starace and Greg Brockman and Hadi Salman and Haiming Bao and Haitang Hu and Hannah Wong and Haoyu Wang and Heather Schmidt and Heather Whitney and Heewoo Jun and Hendrik Kirchner and Henrique Ponde de Oliveira Pinto and Hongyu Ren and Huiwen Chang and Hyung Won Chung and Ian Kivlichan and Ian O'Connell and Ian O'Connell and Ian Osband and Ian Silber and Ian Sohl and Ibrahim Okuyucu and Ikai Lan and Ilya Kostrikov and Ilya Sutskever and Ingmar Kanitscheider and Ishaan Gulrajani and Jacob Coxon and Jacob Menick and Jakub Pachocki and James Aung and James Betker and James Crooks and James Lennon and Jamie Kiros and Jan Leike and Jane Park and Jason Kwon and Jason Phang and Jason Teplitz and Jason Wei and Jason Wolfe and Jay Chen and Jeff Harris and Jenia Varavva and Jessica Gan Lee and Jessica Shieh and Ji Lin and Jiahui Yu and Jiayi Weng and Jie Tang and Jieqi Yu and Joanne Jang and Joaquin Quinonero Candela and Joe Beutler and Joe Landers and Joel Parish and Johannes Heidecke and John Schulman and Jonathan Lachman and Jonathan McKay and Jonathan Uesato and Jonathan Ward and Jong Wook Kim and Joost Huizinga and Jordan Sitkin and Jos Kraaijeveld and Josh Gross and Josh Kaplan and Josh Snyder and Joshua Achiam and Joy Jiao and Joyce Lee and Juntang Zhuang and Justyn Harriman and Kai Fricke and Kai Hayashi and Karan Singhal and Katy Shi and Kavin Karthik and Kayla Wood and Kendra Rimbach and Kenny Hsu and Kenny Nguyen and Keren Gu-Lemberg and Kevin Button and Kevin Liu and Kiel Howe and Krithika Muthukumar and Kyle Luther and Lama Ahmad and Larry Kai and Lauren Itow and Lauren Workman and Leher Pathak and Leo Chen and Li Jing and Lia Guy and Liam Fedus and Liang Zhou and Lien Mamitsuka and Lilian Weng and Lindsay McCallum and Lindsey Held and Long Ouyang and Louis Feuvrier and Lu Zhang and Lukas Kondraciuk and Lukasz Kaiser and Luke Hewitt and Luke Metz and Lyric Doshi and Mada Aflak and Maddie Simens and Madelaine Boyd and Madeleine Thompson and Marat Dukhan and Mark Chen and Mark Gray and Mark Hudnall and Marvin Zhang and Marwan Aljubeh and Mateusz Litwin and Matthew Zeng and Max Johnson and Maya Shetty and Mayank Gupta and Meghan Shah and Mehmet Yatbaz and Meng Jia Yang and Mengchao Zhong and Mia Glaese and Mianna Chen and Michael Janner and Michael Lampe and Michael Petrov and Michael Wu and Michele Wang and Michelle Fradin and Michelle Pokrass and Miguel Castro and Miguel Oom Temudo de Castro and Mikhail Pavlov and Miles Brundage and Miles Wang and Minal Khan and Mira Murati and Mo Bavarian and Molly Lin and Murat Yesildal and Nacho Soto and Natalia Gimelshein and Natalie Cone and Natalie Staudacher and Natalie Summers and Natan LaFontaine and Neil Chowdhury and Nick Ryder and Nick Stathas and Nick Turley and Nik Tezak and Niko Felix and Nithanth Kudige and Nitish Keskar and Noah Deutsch and Noel Bundick and Nora Puckett and Ofir Nachum and Ola Okelola and Oleg Boiko and Oleg Murk and Oliver Jaffe and Olivia Watkins and Olivier Godement and Owen Campbell-Moore and Patrick Chao and Paul McMillan and Pavel Belov and Peng Su and Peter Bak and Peter Bakkum and Peter Deng and Peter Dolan and Peter Hoeschele and Peter Welinder and Phil Tillet and Philip Pronin and Philippe Tillet and Prafulla Dhariwal and Qiming Yuan and Rachel Dias and Rachel Lim and Rahul Arora and Rajan Troll and Randall Lin and Rapha Gontijo Lopes and Raul Puri and Reah Miyara and Reimar Leike and Renaud Gaubert and Reza Zamani and Ricky Wang and Rob Donnelly and Rob Honsby and Rocky Smith and Rohan Sahai and Rohit Ramchandani and Romain Huet and Rory Carmichael and Rowan Zellers and Roy Chen and Ruby Chen and Ruslan Nigmatullin and Ryan Cheu and Saachi Jain and Sam Altman and Sam Schoenholz and Sam Toizer and Samuel Miserendino and Sandhini Agarwal and Sara Culver and Scott Ethersmith and Scott Gray and Sean Grove and Sean Metzger and Shamez Hermani and Shantanu Jain and Shengjia Zhao and Sherwin Wu and Shino Jomoto and Shirong Wu and Shuaiqi and Xia and Sonia Phene and Spencer Papay and Srinivas Narayanan and Steve Coffey and Steve Lee and Stewart Hall and Suchir Balaji and Tal Broda and Tal Stramer and Tao Xu and Tarun Gogineni and Taya Christianson and Ted Sanders and Tejal Patwardhan and Thomas Cunninghman and Thomas Degry and Thomas Dimson and Thomas Raoux and Thomas Shadwell and Tianhao Zheng and Todd Underwood and Todor Markov and Toki Sherbakov and Tom Rubin and Tom Stasi and Tomer Kaftan and Tristan Heywood and Troy Peterson and Tyce Walters and Tyna Eloundou and Valerie Qi and Veit Moeller and Vinnie Monaco and Vishal Kuo and Vlad Fomenko and Wayne Chang and Weiyi Zheng and Wenda Zhou and Wesam Manassra and Will Sheu and Wojciech Zaremba and Yash Patil and Yilei Qian and Yongjik Kim and Youlong Cheng and Yu Zhang and Yuchen He and Yuchen Zhang and Yujia Jin and Yunxing Dai and Yury Malkov},
      year={2024},
      eprint={2410.21276},
      archivePrefix={arXiv},
      primaryClass={cs.CL},
      url={https://arxiv.org/abs/2410.21276}, 
}

@techreport{google2025gemini3pro,
  author      = {{Google DeepMind}},
  title       = {Gemini 3 {Pro} Model Card},
  institution = {Google DeepMind},
  year        = {2025},
  month       = {December},
  url         = {https://storage.googleapis.com/deepmind-media/Model-Cards/Gemini-3-Pro-Model-Card.pdf},
  note        = {Model released November 2025}
}

@article{zhong2025survey,
  title={A survey on vision-language-action models: An action tokenization perspective},
  author={Zhong, Yifan and Bai, Fengshuo and Cai, Shaofei and Huang, Xuchuan and Chen, Zhang and Zhang, Xiaowei and Wang, Yuanfei and Guo, Shaoyang and Guan, Tianrui and Lui, Ka Nam and others},
  journal={arXiv preprint arXiv:2507.01925},
  year={2025}
}

@article{chi2025diffusion,
  title={Diffusion policy: Visuomotor policy learning via action diffusion},
  author={Chi, Cheng and Xu, Zhenjia and Feng, Siyuan and Cousineau, Eric and Du, Yilun and Burchfiel, Benjamin and Tedrake, Russ and Song, Shuran},
  journal={The International Journal of Robotics Research},
  volume={44},
  number={10-11},
  pages={1684--1704},
  year={2025},
  publisher={Sage Publications Sage UK: London, England}
}

@article{team2024octo,
  title={Octo: An open-source generalist robot policy},
  author={Team, Octo Model and Ghosh, Dibya and Walke, Homer and Pertsch, Karl and Black, Kevin and Mees, Oier and Dasari, Sudeep and Hejna, Joey and Kreiman, Tobias and Xu, Charles and others},
  journal={arXiv preprint arXiv:2405.12213},
  year={2024}
}

@article{reuss2025flower,
  title={Flower: Democratizing generalist robot policies with efficient vision-language-action flow policies},
  author={Reuss, Moritz and Zhou, Hongyi and R{\"u}hle, Marcel and Ya{\u{g}}murlu, {\"O}mer Erdin{\c{c}} and Otto, Fabian and Lioutikov, Rudolf},
  journal={arXiv preprint arXiv:2509.04996},
  year={2025}
}

@inproceedings{yue2024mmmu,
  title={Mmmu: A massive multi-discipline multimodal understanding and reasoning benchmark for expert agi},
  author={Yue, Xiang and Ni, Yuansheng and Zhang, Kai and Zheng, Tianyu and Liu, Ruoqi and Zhang, Ge and Stevens, Samuel and Jiang, Dongfu and Ren, Weiming and Sun, Yuxuan and others},
  booktitle={Proceedings of the IEEE/CVF conference on computer vision and pattern recognition},
  pages={9556--9567},
  year={2024}
}

@article{chen2024we,
  title={Are we on the right way for evaluating large vision-language models?},
  author={Chen, Lin and Li, Jinsong and Dong, Xiaoyi and Zhang, Pan and Zang, Yuhang and Chen, Zehui and Duan, Haodong and Wang, Jiaqi and Qiao, Yu and Lin, Dahua and others},
  journal={Advances in Neural Information Processing Systems},
  volume={37},
  pages={27056--27087},
  year={2024}
}

@inproceedings{li2024mvbench,
  title={Mvbench: A comprehensive multi-modal video understanding benchmark},
  author={Li, Kunchang and Wang, Yali and He, Yinan and Li, Yizhuo and Wang, Yi and Liu, Yi and Wang, Zun and Xu, Jilan and Chen, Guo and Luo, Ping and others},
  booktitle={Proceedings of the IEEE/CVF Conference on Computer Vision and Pattern Recognition},
  pages={22195--22206},
  year={2024}
}

\medskip

\newpage
\appendix

\section{Case studies: QA examples}

\subsection{High-quality QAs generated by the Generator}
\textbf{Example 1.}
\begin{center}
    \includegraphics[width=0.5\textwidth]{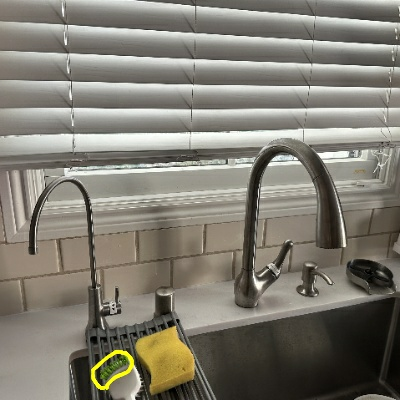}
\end{center}

\vspace{0.5em}

% ==================== ORIGINAL EXAMPLE ====================
\begin{originalbox}[Original Example (GPT)]
\textbf{Question:} What is the likely purpose of the smaller faucet on the left?
\begin{itemize}[leftmargin=2em, itemsep=2pt]
    \item[A.] Dispensing soap
    \item[B.] Providing filtered water
    \item[C.] Spraying water for cleaning
    \item[D.] Controlling water temperature
\end{itemize}
\textbf{Answer:} \texttt{B} \\
\textbf{Reasoning Type:} \textit{Functional Context}
\end{originalbox}

% ==================== JUDGE RESULTS ====================
% \section*{\faGavel~Judge Results}

\begin{geminibox}[\faRobot~Gemini Judge]
\scorebadge{geminicolor}{10}

\vspace{0.5em}
\textbf{Explanation:} The question is grounded, clear, and requires functional reasoning about a specific object in the scene. The visual evidence (a thin, secondary faucet distinct from the main mixer and soap pump) strongly supports the ground truth (filtered water) over the other distractors.
\end{geminibox}

\textbf{Example 2.}
% 77_1

% --- Color Definitions ---
\definecolor{originalcolor}{RGB}{52, 152, 219}   % Blue
\definecolor{geminicolor}{RGB}{155, 89, 182}     % Purple
\definecolor{qwencolor}{RGB}{230, 126, 34}       % Orange
\definecolor{revisedcolor}{RGB}{39, 174, 96}     % Green
\definecolor{finalcolor}{RGB}{231, 76, 60}       % Red
\definecolor{lightgray}{RGB}{245, 245, 245}

% --- Hyperref Setup ---
\hypersetup{
    colorlinks=true,
    linkcolor=blue,
    urlcolor=blue
}

\begin{center}
    \includegraphics[width=0.5\textwidth]{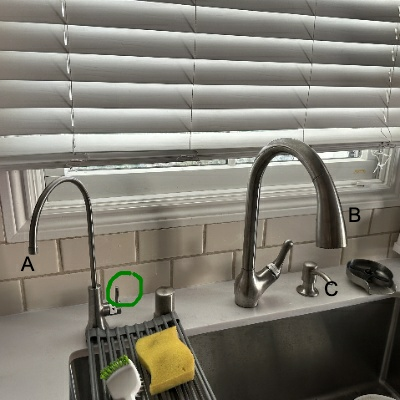}
\end{center}

\vspace{0.5em}

% ==================== ORIGINAL EXAMPLE ====================
\begin{originalbox}[Original Example (GPT)]
\textbf{Question:} Which object is likely used for filtering water?
\begin{itemize}[leftmargin=2em, itemsep=2pt]
    \item[A.] Faucet A
    \item[B.] Faucet B
    \item[C.] Handle C
    \item[D.] Sponge
\end{itemize}
\textbf{Answer:} \texttt{A} \\
\textbf{Reasoning Type:} \textit{Object Reasoning}
\end{originalbox}

% ==================== JUDGE RESULTS ====================
% \section*{\faGavel~Judge Results}

\begin{geminibox}[\faRobot~Gemini Judge]
\scorebadge{geminicolor}{10}

\vspace{0.5em}
\textbf{Explanation:} The question is excellent. It requires visual recognition of specific kitchen fixtures and applies common-sense reasoning regarding their function (distinguishing a main faucet from a secondary filtered water faucet based on size and placement). The options are clear, mutually exclusive, and the ground truth is logically and visually correct.
\end{geminibox}

\textbf{Example 3.}
\begin{center}
    \includegraphics[width=0.5\textwidth]{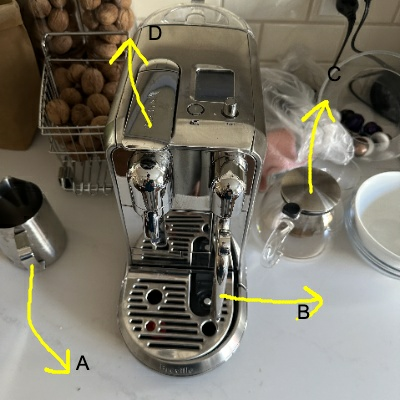}
\end{center}

\vspace{0.5em}

% ==================== ORIGINAL EXAMPLE ====================
\begin{originalbox}[Original Example (GPT)]
\textbf{Question:} Which arrow points to the part of the machine used to froth milk?
\begin{itemize}[leftmargin=2em, itemsep=2pt]
    \item[A.] Arrow A
    \item[B.] Arrow B
    \item[C.] Arrow C
    \item[D.] Arrow D
\end{itemize}
\textbf{Answer:} \texttt{B} \\
\textbf{Reasoning Type:} \textit{Object Reasoning}
\end{originalbox}

% ==================== JUDGE RESULTS ====================
% \section*{\faGavel~Judge Results}

\begin{geminibox}[\faRobot~Gemini Judge]
\scorebadge{geminicolor}{10}

\vspace{0.5em}
\textbf{Explanation:} The QA pair is of high quality. The question asks to identify a specific functional part of the machine, and the choices perfectly align with the visual annotations in the image. The ground truth correctly identifies Arrow B as pointing to the steam wand, which is the part used to froth milk. The distractors are visually distinct and serve their purpose well (e.g., Arrow A points to the milk jug, which is related to milk but is not a part of the machine).
\end{geminibox}

\textbf{Example 4.}
\begin{center}
    \includegraphics[width=0.5\textwidth]{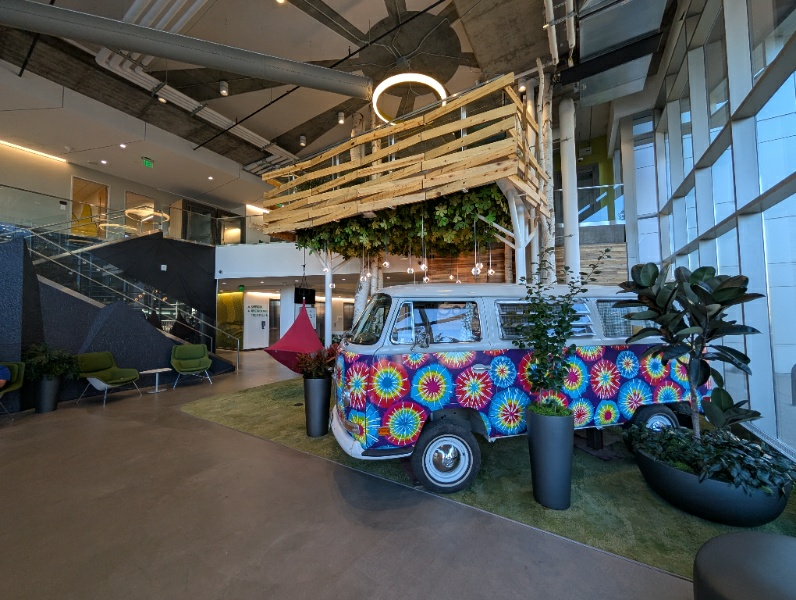}
\end{center}

\vspace{0.5em}

% ==================== ORIGINAL EXAMPLE ====================
%\begin{originalbox}[\faFileAlt~Original Example (GPT)]
\begin{originalbox}[Original Example (GPT)]
\textbf{Question:} What is the likely purpose of the area where the van is placed?
\begin{itemize}[leftmargin=2em, itemsep=2pt]
    \item[A.] A parking lot
    \item[B.] A recreational or lounge area
    \item[C.] A storage area
    \item[D.] A workshop
\end{itemize}
\textbf{Answer:} \texttt{B} \\
\textbf{Reasoning Type:} \textit{Layout Understanding}
\end{originalbox}

% ==================== JUDGE RESULTS ====================
% \section*{\faGavel~Judge Results}

\begin{geminibox}[\faRobot~Gemini Judge]
\scorebadge{geminicolor}{10}

\vspace{0.5em}
\textbf{Explanation:} The question effectively tests high-level scene understanding and contextual reasoning. The options are clear and mutually exclusive, and the ground truth correctly identifies the space as a lounge area based on visual cues like seating, indoor placement, and decorative elements, successfully challenging the prior bias that a vehicle belongs in a parking lot.
\end{geminibox}

\textbf{Example 5.}
\begin{center}
    \includegraphics[width=0.5\textwidth]{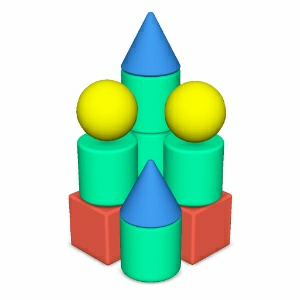}
\end{center}

\vspace{0.5em}

% ==================== ORIGINAL EXAMPLE ====================
%\begin{originalbox}[\faFileAlt~Original Example (GPT)]
\begin{originalbox}[Original Example (GPT)]
\textbf{Question:} What is the correct sequence to build the structure from the base up?
\begin{itemize}[leftmargin=2em, itemsep=2pt]
    \item[A.] Cubes, cylinders, cones, spheres
    \item[B.] Cubes, spheres, cylinders, cones
    \item[C.] Cubes, cylinders, spheres, cones
    \item[D.] Cylinders, cubes, spheres, cones
\end{itemize}
\textbf{Answer:} \texttt{C} \\
\textbf{Reasoning Type:} \textit{Procedural Reasoning}
\end{originalbox}

% ==================== JUDGE RESULTS ====================
% \section*{\faGavel~Judge Results}

\begin{geminibox}[\faRobot~Gemini Judge]
\scorebadge{geminicolor}{8}

\vspace{0.5em}
\textbf{Explanation:} The question is visually grounded and requires valid embodied reasoning about support relationships (e.g., cubes must be placed before the cylinders that sit on them). The ground truth correctly identifies a valid construction sequence. However, there is a minor ambiguity between options A and C. Both are physically valid sequences because the spheres and cones are both 'top-level' objects on their respective stacks and do not depend on each other. Option C is likely the intended answer because the central cone appears to be the highest point of the structure (implying it is placed last), but this relies on a height-based heuristic that isn't strictly defined in the question.
\end{geminibox}

\subsection{Before and after QA revision}
We present illustrative examples in which GPT-4o serves as both the Generator and Reviser, while Gemini 3 Pro acts as the Judge. The following original QA generated by GPT-4o received a score of 3 from the judge. After the first round of revision, the score of this question increased to 10.
\begin{center}
    \includegraphics[width=0.5\textwidth]{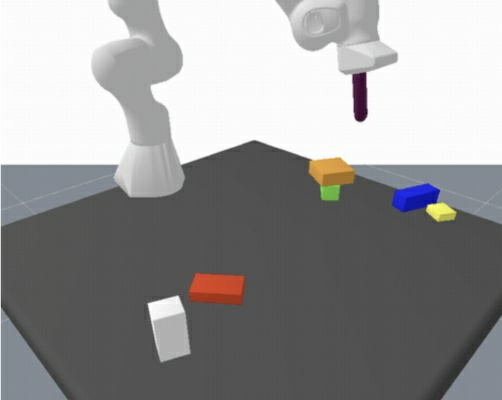}
    % \captionof{figure}{Source image: \texttt{./data/images/77\_0.png}}
\end{center}

\vspace{0.5em}

% ==================== ORIGINAL EXAMPLE ====================
\begin{originalbox}[Original Example (GPT)]
\textbf{Question:} Which block is positioned furthest from the robotic arm?
\begin{itemize}[leftmargin=2em, itemsep=2pt]
    \item[A.] White
    \item[B.] Red
    \item[C.] Blue
    \item[D.] Yellow
\end{itemize}
\textbf{Answer:} \texttt{B} \\
\textbf{Reasoning Type:} \textit{Spatial Reasoning}
\end{originalbox}

% ==================== JUDGE RESULTS ====================
% \section*{\faGavel~Judge Results}

\begin{geminibox}[\faRobot~Gemini Judge]
\scorebadge{geminicolor}{3}

\vspace{0.5em}
\textbf{Explanation:} The ground truth is spatially incorrect. The image shows
a \textbf{white} block in the immediate foreground (bottom-left), which is
physically further from the robot's base (located in the background-left)
than the red block. The red block is positioned \emph{between} the white
block and the robot. Therefore, the answer should likely be White
(or potentially the blocks on the far right depending on exact dimensions),
but certainly not Red.
\end{geminibox}

% ==================== REVISED EXAMPLES ====================
% \section*{\faSyncAlt~Revised Examples}

%\begin{revisedbox}[\faPenAlt~Revised Example (GPT)]
\begin{revisedbox}[Revised Example (GPT)]
\textbf{Question:} Which block is positioned closest to the bottom-left
corner of the platform?
\begin{itemize}[leftmargin=2em, itemsep=2pt]
    \item[A.] White
    \item[B.] Red
    \item[C.] Blue
    \item[D.] Yellow
\end{itemize}
\textbf{Answer:} \texttt{A} \\
\textbf{Reasoning Type:} \textit{Spatial Reasoning}
\end{revisedbox}

\vspace{0.5em}

\begin{geminibox}[\faRobot~Gemini Judge (round 2)]
\scorebadge{geminicolor}{10}

\vspace{0.5em}
\textbf{Explanation:} The question is clear, the options are distinct,
and the ground truth is perfectly aligned with the visual evidence.
The white block is unambiguously the closest to the bottom-left
corner of the platform.
\end{geminibox}

A second example demonstrates a case in which the revision process fails due to images being presented out of chronological order as a result of human intervention.
\begin{figure}[h]
    \centering
    \begin{subfigure}{0.24\textwidth}
        \includegraphics[width=\linewidth]{figures/supplementary/101_0.png}
    \end{subfigure}
    \hfill
    \begin{subfigure}{0.24\textwidth}
        \includegraphics[width=\linewidth]{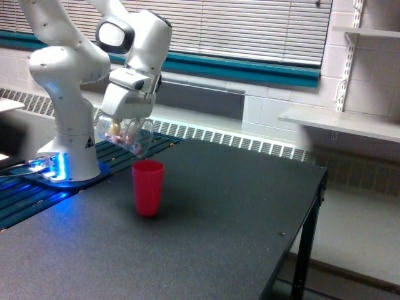}
    \end{subfigure}
    \hfill
    \begin{subfigure}{0.24\textwidth}
        \includegraphics[width=\linewidth]{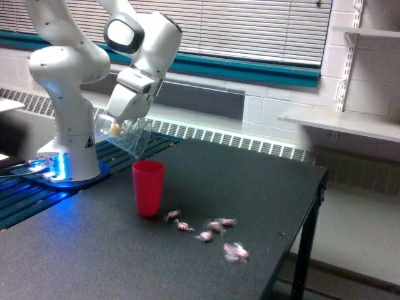}
    \end{subfigure}
    \hfill
    \begin{subfigure}{0.24\textwidth}
        \includegraphics[width=\linewidth]{figures/supplementary/101_3.png}
    \end{subfigure}
\end{figure}

\vspace{0.5em}

% ==================== ORIGINAL EXAMPLE ====================
\begin{originalbox}[Original Example (GPT)]
\textbf{Question:} What happens to the candies over time in the sequence?
\begin{itemize}[leftmargin=2em, itemsep=2pt]
    \item[A.] They are placed in the cup
    \item[B.] They are scattered on the table
    \item[C.] They remain in the robot's hand
    \item[D.] They disappear
\end{itemize}
\textbf{Answer:} \texttt{B} \\
\textbf{Reasoning Type:} \textit{Temporal Reasoning}
\end{originalbox}

% ==================== JUDGE RESULTS ====================
% \section*{\faGavel~Judge Results}

\begin{geminibox}[\faRobot~Gemini Judge]
\scorebadge{geminicolor}{3}

\vspace{0.5em}
\textbf{Explanation:} The images are presented out of chronological order (e.g., pouring in frame 1, candies on the table in frame 3, but an empty table in frame 4). This jumbled sequence makes the question about what happens 'over time' confusing, and the ground-truth answer contradicts the final frame of the provided sequence.
\end{geminibox}

% ==================== REVISED EXAMPLES ====================
% \section*{\faSyncAlt~Revised Examples}

\begin{revisedbox}[Revised Example (GPT)]
\textbf{Question:} What is the final state of the candies in the sequence?
\begin{itemize}[leftmargin=2em, itemsep=2pt]
    \item[A.] They are placed in the cup
    \item[B.] They are scattered on the table
    \item[C.] They remain in the robot's hand
    \item[D.] They disappear
\end{itemize}
\textbf{Answer:} \texttt{B} \\
\textbf{Reasoning Type:} \textit{Temporal Reasoning}
\end{revisedbox}

\vspace{0.5em}

\begin{geminibox}[\faRobot~Gemini Judge (round 2)]
\scorebadge{geminicolor}{3}

\vspace{0.5em}
\textbf{Explanation:} The images are presented out of chronological order. The final frame (Image 4) shows the candies inside the clear cup with an empty table, likely representing the initial state before pouring. The ground-truth answer claims the candies are scattered on the table in the final frame, which directly contradicts the visual evidence of the provided sequence.
\end{geminibox}

\begin{revisedbox}[Revised Example (GPT)]
%\begin{revisedbox}[\faPenAlt~Revised Example (GPT)]
\textbf{Question:} What is the initial state of the candies before any action is taken?
\begin{itemize}[leftmargin=2em, itemsep=2pt]
    \item[A.] The candies are in the cup.
    \item[B.] The candies are scattered on the table.
    \item[C.] The candies are in the robot's hand.
    \item[D.] The candies are not visible.
\end{itemize}
\textbf{Answer:} \texttt{A} \\
\textbf{Reasoning Type:} \textit{Temporal Reasoning}
\end{revisedbox}

\vspace{0.5em}

\begin{geminibox}[\faRobot~Gemini Judge (round 3)]
\scorebadge{geminicolor}{3}

\vspace{0.5em}
\textbf{Explanation:} The images are presented out of chronological order, with the true initial state (candies in the clear cup before pouring) shown in the final frame (Image 4). The ground-truth explanation incorrectly refers to the 'initial frame' as showing this state, which contradicts the provided visual sequence where Image 1 already shows the pouring action in progress.
\end{geminibox}

\section{Failure cases of VLMs}
\textbf{Example 1}: Qwen3-VL-8B, Qwen3-VL-32B-Thinking and Robobrain2.5-8B all fail.
% 77_1

% --- Color Definitions ---

% --- Document ---

% ==================== IMAGE ====================
\begin{center}
    \includegraphics[width=0.5\textwidth]{figures/supplementary/6_0.png}
\end{center}

\vspace{0.5em}

% ==================== QA EXAMPLE ====================
\begin{originalbox}[QA Example (GPT)]
\textbf{Question:} From the robot's perspective, which object is closest to its left side?
\begin{itemize}[leftmargin=2em, itemsep=2pt]
    \item[A.] The can
    \item[B.] The apple
    \item[C.] The orange
    \item[D.] The cup
\end{itemize}
\textbf{Answer:} \texttt{D} \\
\textbf{Reasoning Type:} \textit{Spatial Reasoning}
\end{originalbox}

% ==================== JUDGE RESULTS ====================
% \section*{\faGavel~Judge Results}

\begin{geminibox}[\faRobot~Gemini Judge]
\scorebadge{geminicolor}{10}

\vspace{0.5em}
\textbf{Explanation:} The QA pair is of high quality and correctly tests egocentric spatial reasoning. The question explicitly asks for the object closest to the robot's left side from its perspective. Since the robot is facing the camera, its anatomical left corresponds to the right side of the image. The ground truth correctly identifies the cup, which is on the viewer's right, as the correct answer. The options are clear, mutually exclusive, and the spatial logic is perfectly sound.
\end{geminibox}

% ==================== VLM responses ====================
% \section*{\faSyncAlt~Revised Examples}

\begin{finalbox}[VLM responses]
\textbf{Qwen3-VL-8B:} A (extracted: A) \\
\textbf{Qwen3-VL-32B:} <think>
Got it, let's see. The robot is in the center, and we need to find which object is closest to its left side. Looking at the image: the can is on the left side of the robot, then the apple is being held, the orange is to the right of the apple, and the cup is on the far right. So from the robot's perspective, the left side would have the can. Let's check the options. A is the can. So the answer should be A.
</think>

A (extracted: A) \\
\textbf{RoboBrain2.5-8B:} A (extracted: A)
\end{finalbox}

\textbf{Example 2}.
\begin{center}
    \includegraphics[width=0.5\textwidth]{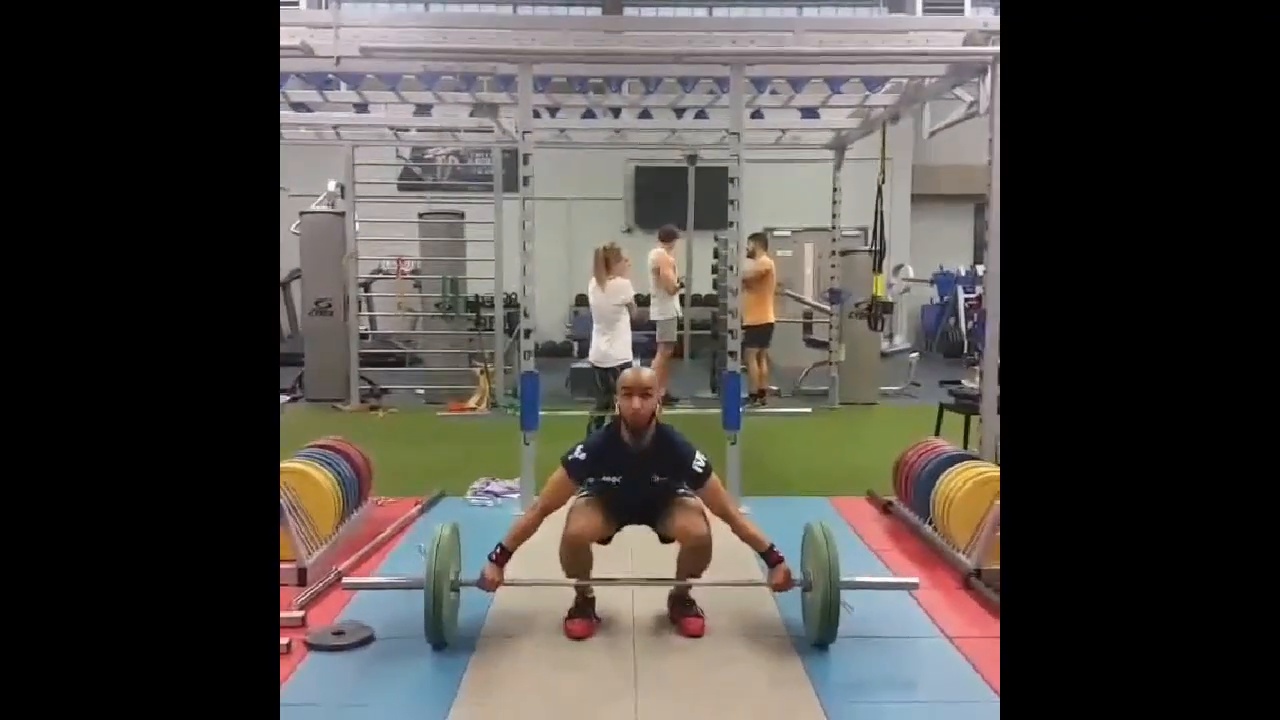}
\end{center}

\vspace{0.5em}

% ==================== QA EXAMPLE ====================
\begin{originalbox}[QA Example (GPT)]
\textbf{Question:} What action is the person in the foreground performing?
\begin{itemize}[leftmargin=2em, itemsep=2pt]
    \item[A.] Squatting
    \item[B.] Deadlifting
    \item[C.] Snatching
    \item[D.] Bench pressing
\end{itemize}
\textbf{Answer:} \texttt{C} \\
\textbf{Reasoning Type:} \textit{Action Recognition}
\end{originalbox}

% ==================== JUDGE RESULTS ====================
% \section*{\faGavel~Judge Results}

\begin{geminibox}[\faRobot~Gemini Judge]
\scorebadge{geminicolor}{9}

\vspace{0.5em}
\textbf{Explanation:} The QA correctly identifies the action based on the visual cues in the static frame. The wide grip on the barbell and the starting posture strongly indicate the setup for a snatch rather than a deadlift or standard squat. The options are clear, mutually exclusive, and the ground truth is accurate.
\end{geminibox}

% ==================== VLM responses ====================
% \section*{\faSyncAlt~Revised Examples}

%\begin{finalbox}[\faPenAlt~VLM responses]
\begin{finalbox}[VLM responses]
\textbf{Qwen3-VL-8B:} B (extracted: B) \\
\textbf{Qwen3-VL-32B:} <think>
Got it, let's see. The person in the foreground is holding a barbell with weights, and their posture—knees bent, back straight, hands gripping the bar. Deadlifting involves lifting the bar from the ground to the hips, so that's the action here. Squatting is more about going down and up with the bar on the shoulders, but here the bar is on the ground. Snatching is a different Olympic lift, and bench pressing is on a bench. So the correct answer should be B.
</think>

B (extracted: B) \\
\textbf{RoboBrain2.5-8B:} B (extracted: B)
\end{finalbox}

\section{Prompts used in dataset generation}

\subsection{Generator prompt}
\begin{lstlisting}[
    basicstyle=\small\ttfamily,
    breaklines=true,
    frame=single]
You are a specialist in designing visual reasoning datasets for embodied robotics. Based on the image(s) captured from the robotic gripper's main or side camera, create high-quality multiple-choice questions that demand grounded visual reasoning. For each question, provide plausible answer options and a single correct answer supported by the visual evidence.
You will create 2-5 multiple-choice questions based on the image(s), focusing on the following reasoning types: 
{
    "Perceptual_Reasoning": {
        "Definition": "Understanding what is seen identifying objects, spatial relations, and temporal dynamics in the environment.",
        "Subtypes": {
        "Object_Reasoning": "Recognizing and reasoning about objects and their physical or functional properties.",
        "Spatial_Reasoning": "Understanding geometric and positional relations among entities in the scene.",
        "Temporal_Reasoning": "Reasoning over time understanding event sequences, motion, and causal effects."
        }
    },
    "Action_Centric_Reasoning": {
        "Definition": "Inferring actions, goals, and task procedures from observed or planned behaviors.",
        "Subtypes": {
        "Action_Recognition": "Identifying manipulation or motion actions performed by humans or robots.",
        "Intention_Reasoning": "Inferring underlying goals or future actions from current behavior.",
        "Procedural_Reasoning": "Understanding multi-step tasks and determining correct action sequences."
        }
    },
    "Social_and_Interaction_Reasoning": {
        "Definition": "Interpreting human roles, attention, and cooperative intentions during interaction.",
        "Subtypes": {
        "Attention_and_Role_Understanding": "Inferring human focus or leadership roles from gaze and gestures."
    },
    "Navigation_and_Environmental_Reasoning": {
        "Definition": "Understanding spatial layouts, obstacles, and paths for safe movement and interaction.",
        "Subtypes": {
        "Layout_Understanding": "Mapping and reasoning about spatial structure or object-scene relationships.",
        "Path_Planning": "Determining feasible routes and avoiding collisions in navigation."
        }
    },
    "Contextual_and_Commonsense_Reasoning": {
        "Definition": "Using functional, environmental, and commonsense knowledge to interpret scenes or predict plausible events.",
        "Subtypes": {
        "Functional_Context": "Inferring the functional purpose of areas or objects in context.",
        "Plausible_Event_Prediction": "Reasoning about likely outcomes or cause-effect relations.",
        "World_Knowledge_Grounding": "Applying commonsense knowledge to interpret visual situations."
        }
    }
}
The sequence of questions should reflect the causal dependencies among these reasoning types:
{
    "Perceptual_Reasoning": {
        "Supports": ["Action_Centric_Reasoning", "Navigation_and_Environmental_Reasoning", "Social_and_Interaction_Reasoning", "Contextual_and_Commonsense_Reasoning"],
        "Depends_On": []
    },
    "Action_Centric_Reasoning": {
        "Supports": ["Procedural_Reasoning", "Contextual_and_Commonsense_Reasoning"],
        "Depends_On": ["Perceptual_Reasoning", "Social_and_Interaction_Reasoning", "Navigation_and_Environmental_Reasoning"]
    },
    "Social_and_Interaction_Reasoning": {
        "Supports": ["Action_Centric_Reasoning", "Contextual_and_Commonsense_Reasoning"],
        "Depends_On": ["Perceptual_Reasoning"]
    },
    "Navigation_and_Environmental_Reasoning": {
        "Supports": ["Action_Centric_Reasoning"],
        "Depends_On": ["Perceptual_Reasoning", "Contextual_and_Commonsense_Reasoning"]
    },
    "Contextual_and_Commonsense_Reasoning": {
        "Supports": ["Perceptual_Reasoning", "Action_Centric_Reasoning", "Social_and_Interaction_Reasoning", "Navigation_and_Environmental_Reasoning"],
        "Depends_On": []
    }
}
Note that you don't need to cover all reasoning types in every image, but choose the most relevant SUBTYPE. Avoid including or providing any visual cues in the question description.

Each question should have 4 answer options (A, B, C, D) with one correct answer. Ensure the questions require reasoning about the visual content and align with the specified reasoning types and dependencies.
For each question, output in the following JSON format:
{"type": reasoning subtype,
"question": concise natural language question,
"choices": list of 4 options (A-D),
"answer": correct option letter,
"explanation": short reasoning why it's correct}

\end{lstlisting}

\subsection{Judge agent prompt}
\begin{lstlisting}[
    basicstyle=\small\ttfamily,
    breaklines=true,
    frame=single,
    % language=json
    % caption={Prompt}
]
You are an Embodied-Reasoning QA Quality Judge. Your role is to assess the quality of a question-choices-answer set in an embodied robotics context and assign a score. Each QA instance includes image(s), multiple answer options, and a ground-truth answer. Your task is to determine whether the QA is meaningful, visually grounded, logically sound, and labeled correctly. Return the JSON object and nothing else.

Evaluate Dimensions:
Question Quality:
- Is the question grounded in the image?
- Does it reflect accurate visual perception?
- Does it rely on correct spatial relationships (e.g., distance, relative position, orientation)?
- Does it require simple observation or genuine embodied reasoning (causality, prediction, multi-step spatial deduction)?

Option Quality:
- Are the choices mutually exclusive and visually distinguishable?
- Is there exactly one correct option?
- Are any options missing or unclear?

Ground-Truth Answer Validity:
- Verify whether the ground-truth answer matches the visual evidence.

Correctness of Spatial Reasoning (if applicable):
- Does the QA rely on correct reasoning about distances, occlusions, viewpoint, or object relations? Flag issues such as poor depth interpretation or incorrect spatial claims.

Overall Usefulness:
- Is the QA suitable for training or benchmarking embodied-reasoning models?
- Does it encourage grounded, high-level reasoning rather than exploiting dataset artifacts (e.g., "priming" or language biases)?

Provide a score with justification:
9-10: High-quality, grounded, unambiguous, correct, strong training value.
7-8: Good but with minor ambiguity or mild reasoning issues.
5-6: Clear flaws-uncertain spatial logic, weak grounding, or questionable labeling.
3-4: Major issues-ambiguous options, wrong ground truth, major reasoning errors.
0-2: Fundamentally broken; visually incorrect question; unusable.

Output Format:
Provide your evaluation in the following JSON format:
{"score": int, "short_explanation": str, "detected_issue": list[str]}
Return the JSON object and nothing else.
\end{lstlisting}

\subsection{Reviser agent prompt}
\begin{lstlisting}[
    basicstyle=\small\ttfamily,
    breaklines=true,
    frame=single,
    % language=json
    % caption={Prompt}
]
You are a specialist in designing visual reasoning datasets for embodied robotics. Based on the image(s) captured from the robotic gripper's main or side camera, create high-quality multiple-choice questions that demand grounded visual reasoning. For each question, provide plausible answer options and a single correct answer supported by the visual evidence.
The following are the question(s) you have generated previously and the judge's evaluation of it:
{question_and_judge_memory}
Score rubric:
9-10: High-quality, grounded, unambiguous, correct, strong training value.
7-8: Good but with minor ambiguity or mild reasoning issues.
5-6: Clear flaws-uncertain spatial logic, weak grounding, or questionable labeling.
3-4: Major issues-ambiguous options, wrong ground truth, major reasoning errors.
0-2: Fundamentally broken; visually incorrect question; unusable.
Using that evaluation:
Identify weaknesses or ambiguities in the original QA
Revise the question, choices, answer, and explanation to fix those issues
Ensure distractor choices are plausible yet clearly incorrect
Output only the final improved QA in the same structured format.
\end{lstlisting}

\subsection{Qwen generator prompt}
\begin{lstlisting}[
    basicstyle=\small\ttfamily,
    breaklines=true,
    frame=single
]
You are a specialist in designing visual reasoning datasets for embodied robotics. Based on the image(s) captured from the robot's main or side camera, create high-quality multiple-choice questions that demand grounded visual reasoning. For each question, provide plausible answer options and a single correct answer supported by the visual evidence.
You will create 1 multiple-choice questions based on the image(s), focusing on the following reasoning type:
{specified_type}: {description}
Avoid including or providing any visual cues in the question description.

Each question should have 4 answer options (A, B, C, D) with one correct answer. Ensure the questions require reasoning about the visual content and align with the specified reasoning type.
Output in the following JSON format:
[{{"type": reasoning type,
"question": concise natural language question,
"choices": list of 4 options (A-D),
"answer": correct option letter,
"explanation": short reasoning why it's correct}}]
Return ONLY the JSON array without any additional text.
\end{lstlisting}

\section{Human evaluation}
\begin{figure}[t]
    \centering
    \includegraphics[width=1\linewidth]{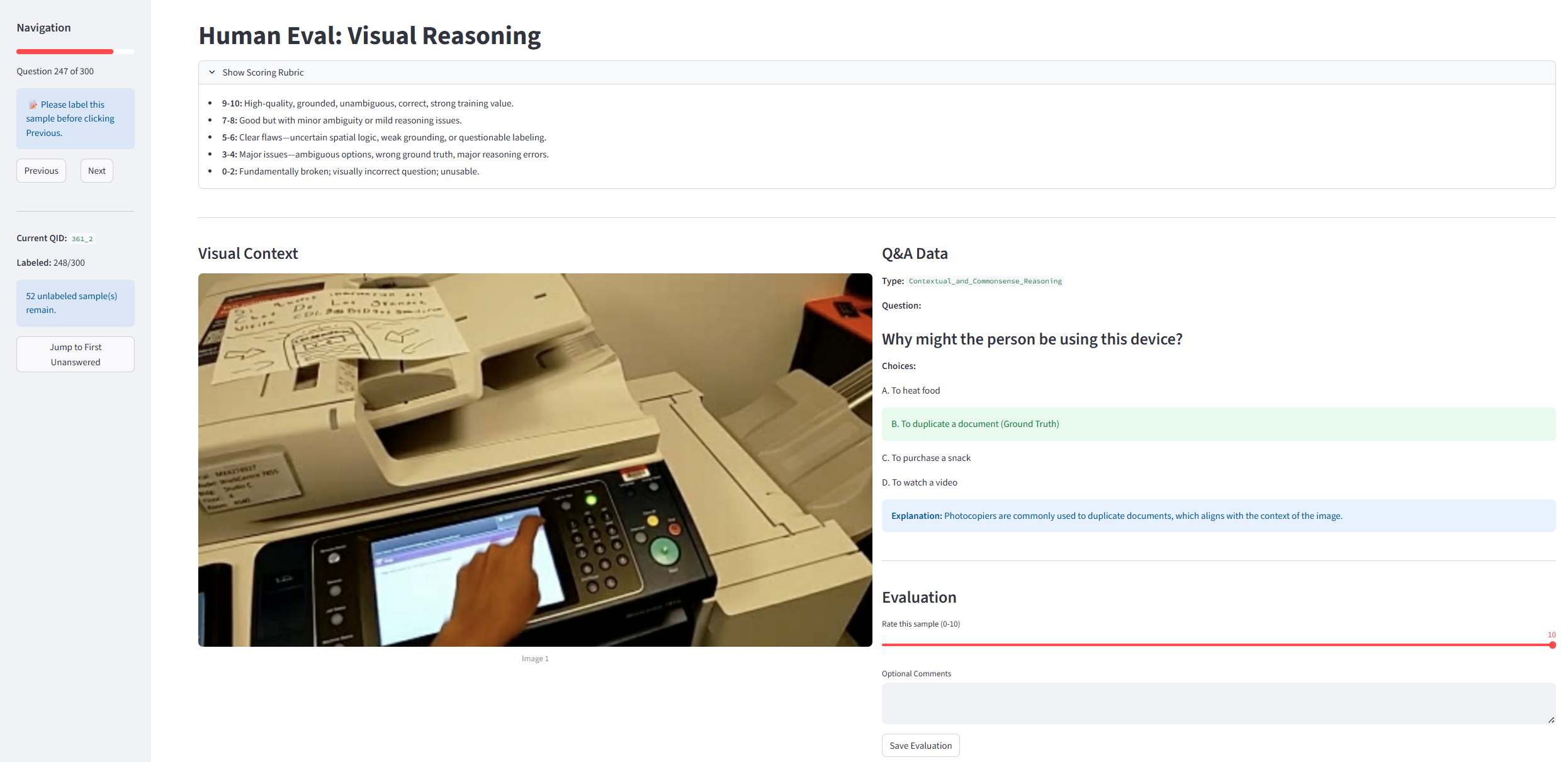}
    \caption{    
    The screenshot of the human evaluation interface.
    }
    \label{fig:human_evaluation}
\end{figure}

A screenshot of the human evaluation interface is shown in Figure~\ref{fig:human_evaluation}.

% Note: Think of the appendix as ``optional reading'' for reviewers. The paper must be able to stand alone without the appendix; for example, adding critical experiments that support the main claims to an appendix is inappropriate. 

\clearpage
% \newpage
%\input{checklist.tex}

\end{document}